\documentclass[12pt, conference]{IEEEtran}

\usepackage[top=0.75in,bottom=0.75in,right=0.75in,left=0.75in]{geometry}
\usepackage[utf8]{inputenc} 
\usepackage[T1]{fontenc}    
\usepackage[hidelinks, breaklinks=true]{hyperref} 
\usepackage{url}            
\usepackage{booktabs}       
\usepackage{amsfonts}       
\usepackage{nicefrac}       
\usepackage{microtype}      
\usepackage{pdfpages}
\usepackage{afterpage}
\usepackage{amsmath,amssymb}
\usepackage[mathscr]{euscript} 
\usepackage{amsthm} 
\usepackage{tikz} 
\usepackage{graphicx} 
\newcommand*{\Scale}[2][4]{\scalebox{#1}{\ensuremath{#2}}} 
\usepackage{framed} 
\usepackage{float} 
\usepackage{multirow}
\usepackage{color}
\usepackage{subfigure}
\usepackage{longtable}
\usepackage{cite}
\usepackage{xcolor}  \definecolor{shadecolor}{rgb}{.95,.95,.95}  
\usepackage[font=footnotesize]{caption}
\usepackage{nicefrac}
\usepackage{dsfont} 
\usepackage[export]{adjustbox} 
\usepackage{cancel} 
\usepackage{wrapfig}

\usepackage[ruled,vlined]{algorithm2e}

\newtheorem{myExample}{Example}

\newtheorem{myRemark}{Remark}

\tikzstyle{every edge}=  [draw]
\tikzstyle{vertex} = [draw,circle,minimum size=1pt]
\tikzstyle{label} = [minimum size=.1pt,font=\scriptsize]
\tikzstyle{title} = [minimum size=.25cm,font=\small]

\newcommand{\bs}[1]{\boldsymbol{#1}}
\newcommand{\bhat}[1]{\boldsymbol{\hat{#1}}}

\newcommand{\gray}{\textcolor{gray!50}}

\def \R{\mathbb{R}}

\def \1{{\mathds{1}}}
\def \T{\mathsf{T}}

\def \spn{{\rm span}}
\def \Ord{\mathscr{O}}
\def \<{\langle}
\def \>{\rangle}

\DeclareMathOperator*{\argmin}{arg\,min}

\def \d{{\rm d}}
\def \r{{\rm r}}

\def \xi{{\hyperref[xiDef]{{\rm x}}}}

\def \M{{\rm M}}
\def \p{{\rm p}}
\def \lambdaa{{\hyperref[lambdaaDef]{\lambda}}}

\def \x{{\bs{\rm x}}}

\def \z{{\hyperref[zDef]{\bs{{\rm z}}}}}

\def \w{\bs{\rm w}}
\def \btheta{{\bs{\theta}}}

\def \U{{\bs{{\rm U}}}}

\def \sU{{\hyperref[sUDef]{\mathbb{U}}}}

\def \i{{{\rm i}}}

\def \k{{\rm k}}

\def \o{{\bs{\omega}}}
\def \ups{{\bs{\upsilon}}}

\def \MSD{{\sc msd}}
\def \RMSD{{R{\sc msd}}}
\def \SRMSD{{SR{\sc msd}}}

\def \PCA{{\hyperref[PCADef]{PCA}}}

\newcommand\Mark[1]{\textsuperscript{#1}}
\ifCLASSINFOpdf
\else
\fi

\begin{document}
\title{GLIMPS: A Greedy Mixed-Integer Approach for \\ Super Robust Matched Subspace Detection}
\author{\IEEEauthorblockN{Md Mahfuzur Rahman\Mark{1}, Daniel Pimentel-Alarc\'on\Mark{2}}
\IEEEauthorblockA{\Mark{1}Georgia State University, \Mark{2}University of Wisconsin-Madison}

}


%


\maketitle

\begin{abstract} 
Due to diverse nature of data acquisition and modern applications, many contemporary problems involve high dimensional datum $\x \in \R^\d$ whose entries often lie in a union of subspaces and the goal is to find out which entries of $\x$ match with a particular subspace $\sU$, classically called \emph {matched subspace detection}. Consequently, entries that match with one subspace are considered as inliers w.r.t the subspace while all other entries are considered as outliers. Proportion of outliers relative to each subspace varies based on the degree of coordinates from subspaces. This problem is a combinatorial NP-hard in nature and has been immensely studied in recent years. Existing approaches can solve the problem when outliers are sparse. However, if outliers are abundant or in other words if $\x$ contains coordinates from a fair amount of subspaces, this problem can't be solved with acceptable accuracy or within a reasonable amount of time. This paper proposes a two-stage approach called \emph{Greedy Linear Integer Mixed Programmed Selector} (GLIMPS) for this abundant-outliers setting, which combines a greedy algorithm and mixed integer formulation and can tolerate over 80\% outliers, outperforming the state-of-the-art.
\end{abstract}

\IEEEpeerreviewmaketitle
\section{Introduction}
\label{introSec}
Approaches for data analysis has been greatly advanced in recent years. It results in the emergence of high-dimensional data in all areas of science, business, and engineering problems. To this end, high-dimensional data can often be best explained using low-dimensional linear subspaces. In this model, a high dimensional datum $\x \in \R^\d$ is a point lying in a subspace $\sU$. The task of testing whether a given datum $\x$ lies in the given subspace $\sU$ or not, is commonly known as {\em matched subspace detection} (\MSD) \cite{msd}. Testing whether $\x$ is a complete inlier vector or not can be readily accomplished using the mathematical relationships between a vector $\x$ and its corresponding subspace $\sU$. The classical formulation of this \MSD\ problem is formulated as a hypothesis test problem for which optimality of the solution is guaranteed based on the amount of relative signal energy of $\x$ w.r.t. the given subspace $\sU$ \cite{msd, scharf}. Different variants of this {\em matched subspace detection} problem has numerous applications in literature including remote sensing systems \cite{targetLocalization, radar}, neural activation detection \cite{medical}, anomaly detection \cite{anomaly, multivariateAnomaly}, and communications \cite{communications}. In those applications, it is assumed that outliers are sparse in datum. However, many contemporary applications and its variants including computer vision \cite{kanatani},  hyperspectral imaging \cite{hyperspectral}, recommender systems \cite{netflix, prize}, lidar \cite{lidar}, and many more inherently may allow vast majority of outliers in problem setting. Furthermore, several other factors (e.g. missingness , inconsistencies, noises and outliers in data) may also come into picture to make the problem very difficult to solve. To cope with these varied situations, a several other modified formulations of \MSD\ \cite{msdAdaptive,msdGaussian,msdSteering} are proposed in the literature. For instance, it may be often the case that full measurement of $\x$ is not feasible or not allowed as assumed in \cite{msdMissing, grouse}.  

In this paper, we will assume two variants of {\MSD}  formulation, namely {\em robust matched subspace detection} (\RMSD) and {\em super robust matched subspace detection} (\SRMSD). By \RMSD, we mean a few number of coordinates of $\x$ ($ < \frac{\d}{2}$) is not just slightly perturbed by small noises. Instead, those entries, by any means, have been replaced completely by wrong values (outliers). \SRMSD\ refers to the case that the most of the coordinates ($>> \frac{\d}{2}$) of the given vector $\x$ has been severely corrupted. In both cases, the goal is to identify the inlier entries in $\x$. 
\section{Related Work}
\label{relatedsec}

The \RMSD\ variant of the classical \MSD\ problem can be easily tackled using the $\ell_1$-norm minimization of the residual $\x - \U\theta$ because the solution always favors majority entries which is the case in \RMSD\ (number of inlier entries $> \frac{\d}{2}$). This formulation is closely related to LASSO \cite{lasso} as it also uses $\ell_1$-loss as regularization parameter. This \RMSD\ setting has been well studied in practice because numerous applications can be modeled as \RMSD\ problems \cite{grasta, motionSegmentation, survey, review, semantic2,  roseta,  apg, r2pca}. However, \RMSD\ frameworks are not easily extensible for the problems to be modeled as \SRMSD.
The \MSD\ or \RMSD\ problem is often approached using randomized methods like \cite{ransac, aransac, usac} but it is not applicable in \SRMSD\ setting because of its combinatorial NP-hard nature. Moreover, the solutions provided by randomized methods are not guaranteed to be globally optimal \cite{IR-LP, GORE, Atree, MaxFS, dpca, exactpenalty, biconvex, multiplestructure} even if it is an \RMSD\ problem. However, a very recent bi-convex programming based method is proposed in \cite{biconvex} that aims to refine the result obtained from RANSAC using a second-stage bisection search in the remaining solution space. However, RANSAC itself is sub-optimal in solving an \SRMSD\ problem. Another variant of RANSAC is recently proposed in \cite{arsac} that ranks measurements in each iteration. Furthermore, it necessitates some parameter configurations that makes the approach harder to implement in practice. A deterministic and locally convergent method proposed in \cite{exactpenalty} can handle a large proportion of outliers but it is slow and heavily dependent on RANSAC  or Least-Squares based initial set of solutions. 
 In quest of global optimality for \SRMSD\ class of problems, several approaches have been proposed in recent years. For example in MaxFS \cite{MaxFS}, similar to \SRMSD\ problem is formulated as a set of infeasible constraints and the goal is to find the maximum feasible subset from the set. It uses deterministic branch-and-bound (BnB) methods to find the solution but for large-scale problems, unfortunately it may take exponential time \cite{Atree}. Moreover, MaxFS only guarantees solution for small-scale homogenous linear systems. However, linear matrix inequality constraints within BnB method is proposed in \cite{LMI} to obtain reduced time requirement and handle large amount of outliers, but it requires some pre-defined structure in the objective variables. Recently, iterative re-weighted $\ell_1$-minimization method proposed in \cite{IR-LP} can find comparatively large maximal inlier set but it requires iteratively solving a number of similar sub-problems causing it slower than RANSAC class of algorithms. In addition, it is not supportive for \SRMSD\ formulation as its performance is highly determined by the proportion of outliers.  Another tree-based technique is recently proposed in \cite{Atree} to solve the problem in reduced amount of time using application dependent heuristics and based on certain pre-assumption that residual structure is to be quasi-convex. Unfortunately, in many situations, these heuristics may not be available and the assumption on the residual structure may not be accurate in practice. 

An MILP-based outlier removal technique, called GORE, is proposed in \cite{GORE} but it can only be used as a preprocessing routine for the class of algorithms solving maximum consensus set problem. However, its removal is guaranteed up to 20\% and can't be worthwhile if outliers are abundant. More recently, a heuristic greedy algorithm is proposed in \cite{erasure} for subspace tracking that aims to find the most probable outlier entry testing whether removal of this entry minimizes the gap between $\x$ and its projection onto $\sU$. It is observed that this greedy algorithm, which runs in polynomial time $\Ord(\d^2)$,  outperforms the method based on $\ell_1$-minimization of the residual and tolerates up to 60\% outliers. 

In this paper, we propose GLIMPS which is a joint effort of the greedy \cite{erasure} and MILP algorithms to utilize the power of a fast (greedy) and an exact (MILP) algorithm to solve an \SRMSD\ problem. We show that GLIMPS can outperform other existing approaches including $\ell_1$-minimization, the greedy and MILP algorithms. We also show that $\ell_1$-minimization even when assisted with the greedy algorithm can't perform very well. 

\section{ Contributions of this paper}
\noindent
The contribution of this paper can be summarized as follows:
\begin{itemize} 
\item GLIMPS offers a two-stage approach that exacts the bests of both greedy and MILP algorithms. To make it more precise, greedy algorithm is fast and its performance in handling outliers is better than other randomized and $\ell_1$-based approaches. However, if outliers are abundant (80\% or more), then in very few cases, it unexpectedly removes some inliers as outliers. This unintended phenomenon makes greedy algorithm failed in majority of trials if $\x$ is severely outlier-dominant. On the other hand,  MILP is an exact algorithm, however because of the combinatorial nature of \SRMSD\ problem, it may take exponentially long time to solve the problem. In contrast, GLIMPS reduces the search space for MILP by removing a certain percentage of most harmful entries using greedy algorithm and thus eventually makes MILP faster. 

\item By our proposed greedy algorithm in the first stage, we are providing means to warm-start MILP and thus scalability of MILP has been noticeably increased in GLIMPS. 

\item Greedy algorithm and MILP can tolerate up to 60\% and 76\% outliers respectively when they are separately used. On the other hand, GLIMPS offers tolerance up to 80\% outliers. 

\item Existing approaches like $\ell_1$-minimization \cite{grouse,grasta, candes-recht, candes-tao, recht} assume subspaces of specific coherence and uniformity of outlier locations in $\x$. In contrast, GLIMPS has no restriction on subspace structure or outlier locations. 

\item \SRMSD\ is a generalized version for a class of problems. So, the idea of GLIMPS can easily be applied for other tasks if we can use subspace based model for the problems. Examples include robust \phantomsection\label{PCADef}\PCA\ (principal component analysis), background separation, subspace tracking \cite{grouse,grasta,roseta,erasure}, and matrix completion \cite{candes-recht, candes-tao, recht, almNIPS, altLRMC, LRMCpimentel, HRMC,  gssc, infoTheoretic, elhamifarNIPS}. More details are discussed in Section \ref{motivationSec}.

\item MILP can be implemented with a few lines of code using high-level mathematical optimization languages like AMPL \cite{ampl} and JUMP \cite{jump}. A variety of solvers like GUROBI \cite{gurobi} and CPLEX \cite{cplex} are readily available for academic uses. 
\end{itemize}

\section{Motivating Applications}
\label{motivationSec}

\MSD\ is the classical setting of \RMSD\ and \SRMSD. Moreover, the setting of \RMSD\ and \SRMSD\ can be used to model many applications as mentioned in Section \ref{introSec} and \ref{relatedsec}. In fact, the idea of \SRMSD\ can be applied when subspace-based modeling is feasible. In this section, some applications of \SRMSD\ in recommender systems, network inference, signal processing, computer vision, and metagenomics are briefly described.

{\bf Recommender Systems.}
Organizations like Netflix, YouTube, Amazon and Facebook, based on their requirements, use recommender systems for video, music, product and content recommendations. Different approaches including k-NN, Pearson correlation coefficient and ensemble methods \cite{prize} have been proposed to solve the problem. However, as the user preference vector $\x$ usually lies in a low dimensional subspace, low-rank matrix completion is considered as potentially the best model whatsoever \cite{candes-recht, candes-tao, recht, almNIPS, altLRMC, LRMCpimentel}. Unfortunately, if a single account is shared by multiple users, then the preference vector $\x$ may contain entries that belong to different subspaces. This situation can easily be modeled as an \SRMSD\ problem due to predominance of outliers with respect to each subspace. 

{\bf Networks Inference.}
According to system biology, understanding a system can be accomplished establishing correlations among its components \cite{biology}. This idea can be applied to understand a broad class of networks (internet, biological networks, social networks, and many more) \cite{network, network7}. In general, these interactions among components can be analyzed using mathematical graph theory, where, for example in biological network, each node represents an entity like DNA, RNA, protein and small molecules. Edges may represent activation levels or weights referring to confidence levels, strength, and reaction speeds. This network can be decomposed into multiple strongly connected components which we may refer to as subnets. In this setting, each node in subnet $\k$, can be encoded as a datum $\x$ which , in turn, can be represented as a linear combination of two levels of interactions: 1) interaction to its own subnet $\k$ and 2) interaction between subnet $\k$ and other subnets. According to this modeling, $\x$ lies in a subspace $\sU$. Because of the dynamic behavior of interactions, the state of each node may change dynamically over time and consequently $\x$ can have measurements from different subspaces that results in majority of outliers with respect to each subspace and hence we can leverage \SRMSD\ algorithms for these class of problems. 

{\bf Signal Processing}
In signal processing, for detection problems, usually a time-series data is given and the task is to identify how the data was actually generated. That is, whether the given vector $\x$ entirely consists of noises or outliers or it lies in some subspace containing the spectral signature of the signal. This {\em matched subspace detector} is the generalization of a special case called {\em matched signal detector} also called the {\em matched filter} which is considered as an important building block of signal processing \cite{msd}. In modern signal processing applications, e.g. target detection or anomaly detection is treated as a binary classifier that labels pixels of input images as pixels of interest or as anomalous pixels. In this detection problem, a fixed background can be treated as a data point lying in a low dimensional linear space and targeted or anomalous pixels can be treated as outliers. Based on the position and orientation of the cameras or receiving stations, it may have different amount of outliers. This idea can also be applied to identify peculiarities in solar images and thus infer about any potential events. 

{\bf Computer Vision.}
Computer vision, as a sub-field of signal processing, often tries to segment background from the foreground to detect events occurring in a video. Though the background in traffic or video surveillance problems is subject to minor variation with environmental or instrumental changes, it can still be modeled as a low-dimensional subspace. For this kind of formulation, each video frame (an image) can be vectorized as $\x$ and background can be approximated as a subspace $\sU$. The task is to find which entries in $\x$ match with $\sU$ (inliers) and which entries don't match (outliers). Depending on the position and orientation of cameras and involving objects, frames i.e., $\x$ can have different number of outliers. If foreground objects are far from cameras, the outliers become sparse in $\x$ as assumed in \RMSD\ , robust \PCA\ \cite{motionSegmentation, survey, roseta, review, apg, r2pca} and other subspace tracking approaches \cite{grouse,roseta,erasure}. However, in some data acquisition situations, this is always not the case and number of outliers can be a dominant majority. It turns out that we need to precisely handle the situation of \SRMSD. This kind of modeling can also be applicable for image inpainting \cite{inpainting}, scene reconstruction, designing industrial robots and many more. 

{\bf Metagenomics.}
To better understand different microorganisms present in an environmental sample, traditionally microbiologists used cultivation-based methods which in fact replicate certain genes to create biodiversity. Unfortunately, this approach is less effective in the sense that it always misses most of the microbial diversity. Recently, metagenomics, a recent field of microbial study, aims to provide means to understand all of the genomes from an environmental DNA sample, which is a mixture of genes from multiple organisms. Each genome is a collection of genes and representative of one organism. So, the task is to find gene-wise identities of genome present in the mixed sample. Metagenomics is supposed to advance knowledge and has vast impact on practical applications in the areas of medicine, agriculture and engineering. For example, in agriculture, better understanding the relationships between plants and microbes can reduce the risk of diseases in crops and livestock, provide enhanced means for adaptive farming \cite{metagenomics6}. Metagenomic problem can easily be modeled as an \SRMSD\ problem where $\x$ refers to the DNA sample collected from the environment, i.e., each gene will correspond to one entry in $\x$. Each genome can be modeled as a subspace and the goal is to find out which gene (coordinate) corresponds to which genome (subspace). As the sample $\x$ contains entries (genes) from multiple organisms, there will be few inliers (correct genes) for each subspace. This formulation matches with our \SRMSD\ setting. 
\section{Problem Formulation}
\label{probSec}
The problem is formulated as follows: consider an $\r$-dimensional subspace \phantomsection\label{sUDef}$\sU$ and a vector $\x$ in the ambient dimension $\R{}^{\d}$. The given vector $\x$ contains some coordinates correct relative to the subspace $\sU$ and the rest are all outliers relative to $\sU$. \SRMSD\ is defined as finding set of indices $\o$ of all the inlier coordinates in $\x$, where $\o$ means the largest $\ups \subset [\d]:=\{1, \dots, \d\}$ such that $\x_\ups$ lies in the subspace $\sU_\ups$. The subscript $\ups$ refers to the restriction to the coordinates indicated in $\ups$. The subspace $\sU$ is spanned by $\U \in \R{}^{\d \times \r}$, whereas $\sU_\ups$ is spanned by $\U_\ups \in \R{}^{|\ups| \times \r}$. 
\begin{myExample}
\label{intuitionEg}
Consider $\d = 5$ and $\r = 2$.
Given that \begin{align*}
\U \ = \ \left[ \begin{matrix}
1 & 0 \\
3 & 2 \\
5 & 4 \\
7 & 6 \\
9 & 8 
\end{matrix} \right],
\hspace{.5cm}
\x \ = \ \left[ \begin{matrix}
4 \\ 14 \\ 0 \\ 34 \\ 44
\end{matrix} \right],
\end{align*}
and according to formulation, $\sU=\spn\{\U\}$. \\ It is evident that $\o=\{2,4,5\}$, because
\begin{align*}
\x_{\o} = \left[ \begin{matrix}
14 \\ 34 \\ 44
\end{matrix} \right]
\ = \
\left[ \begin{matrix}
3 & 2 \\
7 & 6 \\
9 & 8
\end{matrix} \right]
\left[ \begin{matrix}
4 \\ 1
\end{matrix} \right]
\ = \ \U_{\o} \btheta,
\end{align*}
and because $\o$ is the largest $\ups$ for which $\x_\ups$ lies in $\sU_\ups$. 
\end{myExample}

The fact that $\btheta$ we derived in the previous example is correct and $\x_\o$ are really inliers can be easily justified as follows: Given that the dimension of the subspace is $\r = 2$ and consequently, the dimension of the coefficient vector $\btheta$ is $2$. Hence, as we know $\x = \U\btheta$, we have two unknowns in the chosen system of equations $\x_\o = \U_\o\btheta$.  Now, if we choose $m = |\o| \leq \r$ equations, the system is either underdetermined or uniquely determined, and we must get a solution for $\btheta$. For example, if we choose $\o = \{1, 3 \}$, we have a solution $\btheta=[4 \ -5]^\T$. Same is the case if we choose even fewer equations. However, this solution is not guaranteed because the system is under-constrained or uniquely constrained. To guarantee that the set of indices $\o$ and the corresponding inliers are correct, we need to find a solution for an overdetermined system, i.e. we need at least $\r+1$ equations.

\section{Greedy Linear Integer Mixed Programmed Selector}
\label{glimpsSec}
As we have mentioned in Related Work section,  to address the problem in question, i.e. to solve \SRMSD\ problem,  greedy algorithm and mixed-integer linear program (MILP) based formulation have been proposed in the literature. However, if the fraction of outlier becomes huge, these techniques when used separately either fail to give correct results or take exponentially longer time. Our two-stage approach to address \SRMSD\ is a combination of a greedy algorithm and a MILP to improve the existing results. In its first stage, we iteratively remove those coordinates (one per each iteration) whose removal minimizes the gap between $\x$ and its projection onto the subspace $\sU$. Projections are computed dynamically as we restrict coordinates only to the existing coordinates. Then, in the second stage, we apply a MILP to identify the remaining inliers and outliers. Once we identify the remaining inliers, we obtain the coefficient of the inliers of $\x$, $\btheta$ which in turn allows us to identify {\em all} the inliers and outliers in $\x$. 

\medskip
\noindent 
{\bf First Stage: Greedy Algorithm.}
Suppose $\o^c$ is a set of coordinates of the given vector $\x$ that are already found to be outliers. We compute $\d-|\o^c|$ pair (for $\U$ and $\x$) of projections onto each set of the coordinates $\{1, \dots, \d\} \setminus \{\o^c\cup \{i\}\}, \forall i\in \{1, \dots, \d\} \setminus\o^c$. Let $\o_i=\{1, \dots, \d\} \setminus \{\o^c\cup \{i\}\}$.
Mathematically, $\U_{\o_i}=P_{\o_i}\U$ and $\x_{\o_i}=P_{\o_i}\x$, where $P_{\o_i}$ is the projection operator when $\x$ and $\U$ are restricted to the set of coordinates $\o_i$. Then, we compute projection of $\x_{\o_i}$ onto  the projected subspace $\U_{\o_i}$, i.e., $\hat{\x}_{\o_i}=\U_{\o_i}(\U_{\o_i}^{\T}\U_{\o_i})^{-1}\U_{\o_i}^{\T}\x_{\o_i}$. For each $\o_i$ as defined earlier, we compute the ratio of two norms, $r_i=\frac{\|\hat{\x}_{\o_i}\|}{\|\x_{\o_i}\|} \leq 1$. It is known that larger $r_i$ implies $\x_{\o_i}$ and $\hat{\x}_{\o_i}$ are closer. Intuitively, it implies if we remove $i$-th coordinate along with other $\o^c$ coordinates, $\x$ gets closer to the subspace $\sU$ whose basis is $\U$. Because of having a clear set of intuitive choices, this approach is called greedy erasure (outlier removal) algorithm. We identify the coordinate with largest ratio, and the coordinate is removed as an outlier adding the index to the outlier set of indices $\o^c$. Obviously, at the beginning of the algorithm, $\o^c=\emptyset$. We iterate this process until we have sufficient number of inliers compared to the number of outliers so that MILP can succeed in the second stage. The greedy algorithm used in the first stage is shown in the following algorithm. 
\begin{algorithm} 
 \caption{Greedy Algorithm (First Stage)}
\KwIn{Subspace basis $\U$, subspace dimension $\r$, observed vector $\x$, removal percentage $p$}
\KwOut{A reduced version $\x_\ups$ of input vector $\x$, to be solved by MILP}

\Scale[.7]{\gray{1}}. Initialize sets of outliers as $\o^c=\emptyset$\;
\Scale[.7]{\gray{2}}. \For{$i=1$ \KwTo $d$ and $i \not \in \o^c$ } {
    \Scale[.7]{\gray{2.1}}. Project $\x$ and $\U$ onto the existing coordinates except $i$ and obtain $\x_{\o_i}$ and $\U_{\o_i}$\;
     \Scale[.7]{\gray{2.2}}. Project $\x_{\o_i}$ onto $\U_{\o_i}$ and obtain  $\hat{\x}_{\o_i}=\U_{\o_i}(\U_{\o_i}^{\T}\U_{\o_i})^{-1}\U_{\o_i}^{\T}\x_{\o_i}$\;
     \Scale[.7]{\gray{2.3}}. Compute $r_i=\frac{\|\hat{\x}_{\o_i}\|}{\|\x_{\o_i}\|}$\;
   }
  
 \Scale[.7]{\gray{3}}. Find $r_m=max(r_i) , \forall i\in \{1, \dots, d\} \setminus\o^c$\;
\Scale[.7]{\gray{4}}. Remove $m$-th coordinate and update  $\o^c= \o^c\cup \{m\}$\;
\Scale[.7]{\gray{5}}. Repeat steps $2$ \KwTo $4$ until we remove desired number of coordinates and thus obtain $\x_\ups$\;
\end{algorithm}


\noindent
{\bf Second Stage: MILP} As output from the greedy stage, we obtain $\x_\ups$---a reduced version of $\x$. All the entries of $\x_\ups$ are not necessarily inliers. However, it is expected after being partially processed by greedy algorithm in the first stage that outliers in $\x_\ups$ are less dense than in $\x$. We apply  MILP (mixed integer linear program) formulation on $\x_\ups$ which is supposed to perform better as the number of variables in $\x$ is reduced before feeding into MILP. The MILP for $x_\ups$ can be formulated as follows:

\begin{framed}
\begin{align}
\label{mipEq}
\argmin_{\z \in \{0,1\}^{|\ups|}, \atop \btheta \in \R^\r}
\ \|\z\|_2^2 
\hspace{.5cm} \text{s.t.} \hspace{.5cm}
|\x_\ups-\U_\ups\btheta| \leq \M \z,
\end{align}
\end{framed}

In this formulation, $\M$ is a sufficiently big constant useful for MILP formulation \cite{bigM}. We should keep $\M$ as small as possible to avoid any potential numerical instability. Fortunately, we can use problem structure to choose $\M$. For example, we could use \begin{align} \M = \max_{\i \in [\d]} \ |\U_\i\btheta - \x_\i | \end{align} 
\phantomsection\label{zDef}$\z$ is a vector of binary variables and assumes zero entries for inliers. To make it precise, let $\bhat{\z}$ be the solution to \eqref{mipEq}  and          $\bhat{\o}$ be the set of indices in $\bhat{\z}$ such that $\bhat{\z}_{\bhat{\o}} = \bs0$. Because $|\x_{\bhat{\o}} - \U_{\bhat{\o}} \btheta | \leq \M \bhat{\z}_{\bhat{\o}}= \bs{0}$ implies $|\x_{\bhat{\o}} - \U_{\bhat{\o}} \btheta | = 0$. It means all the entries indicated in $\x_{\bhat{\o}}$ match with $\sU_{\bhat{\o}}$ for the $\btheta$ that MILP obtains through its searching process. Moreover, $\M$ and $\z$ (takes $1$ for outlier) collectively satisfy constraints for outliers. That is, the idea is to find a suitable $\btheta$ that makes as many inliers as possible maximizing the number of zeros (largest possible $\o$) in $\z$ or in other words, minimizing $\|\z\|_2^2$. Once we find correct $\btheta$ for $\x_\ups$ with at least $|\o| \geq \r+1$ inliers, then we can find all the inliers in $\x$. 

\medskip
\noindent
{\bf MILP Formulation for Noisy Data} 
As we mentioned earlier, high-dimensional data in different problems can be modeled as points lying in a subspace. However, in real setting, that may always not be the case. That is, there might be small modeling noise causing $\x_\ups$ to be close to $\sU_\ups$ but not exactly lying on it. That is, if $\o$ is the set of inlier indices in $\x_\ups$, then $\x_\o \approx \U_\o \btheta$ holds instead of  $\x_\o = \U_\o \btheta$ in noiseless version. Alternatively, for noise case, we can write $\x_\o = \U_\o \btheta + \w_\o$, where $\w_\o$ refers to the vector of small noises associated with inlier entries. These noises can be taken into account by a minor variation in our noiseless version \eqref{mipEq} of MILP formulation as follows: 

\begin{align}
\label{noiseEq}
\argmin_{\z \in \{0,1\}^{|\ups|}, \atop \btheta \in \R^\r, \ \w \in \R^{|\ups|}}
\ \|\z\|_2^2 + \lambdaa \|\w\|_2^2
\hspace{.5cm} \text{s.t.} \hspace{.5cm}
|\x_\ups-\U_\ups\btheta-\w| \leq \M \z,
\end{align}
where \phantomsection\label{lambdaaDef}$\lambdaa>0$ is a regularization parameter that quantifies how much noise we will allow in our noisy formulation for the entries to be treated as inliers. $\lambdaa = 0$ implies that noise has no restriction on $\w$ in \eqref{noiseEq}. Consequently, all the entries in $\x_\ups$ will be treated as inliers and we will end up with an erroneous $\btheta$. It happens because for any $\btheta$, it is possible to satisfy the constraint in \eqref{noiseEq} with $\w =\x_\ups -\U_\ups \btheta$ and $\z=\bs{0}$. In another extreme, if $\lambdaa = \infty$, it forces $\w$ to be $\bs{0}$ and allows no noise. In general, between these two extremes, if $\lambdaa$ is smaller, the formulation in \eqref{noiseEq} will allow more noise. On the other hand, if $\lambdaa$ is larger, it penalizes large values of $\w$ and thus allows less noise. For our experiments, we use moderately large value for $\lambdaa$. We show in Section \ref{experimentsSec} that GLIMPS tolerates a reasonable amount of noise with almost no compromise with accuracy.

\begin{myRemark}
As we already mentioned, finding $\r+1$ or more inliers is as good as finding {\em all} of the inliers in $\x$. This is because we can estimate true coefficient vector $\btheta$ with only a set of inliers $\o$, $|\o| \geq \r+1$ using  $\btheta=(\U{}_\o^\T \U{}_\o)^{-1} \U{}_\o^\T \x_\o$. As the coefficient vector $\btheta$ has only $\r$ unknowns, and we at least have $\r+1$ equations in the system. Consequently, the system of equations is overdetermined and the solution obtained for $\btheta$ is correct.  Once, we obtain $\btheta$, we can easily obtain the entire correct $\x$ using $\x = \U\btheta$ even if $\x$ is partially observed or if it is severely corrupted by outliers. This idea can easily be applied to complete missing or corrupted data in applications such as background estimation when covered by foreground object or in image inpainting \cite{inpainting} where we want to estimate corrupted or outlier pixels in the image w.r.t the background. 
\end{myRemark}

\section{Experiments}
\label{experimentsSec}
In this section, we study the performance of GLIMPS on synthetic  data as a function of the fraction of outliers $\p$, and the noise variance $\sigma^2$. For comparison, we test GLIMPS against a similar MILP formulation, the very recent greedy {\em Erasure} algorithm in \cite{erasure} and the greedy-assisted $\ell_1$-minimization (combination of greedy and $\ell_1$-minimization), which tolerate larger amounts of outliers than other existing algorithms. In our experiments, we measure accuracy as the normalized error between the true and estimated inlier coefficients $\btheta$ and $\bhat{\btheta}$, i.e., $\nicefrac{\|\btheta - \bhat{\btheta}\|_2}{\left ( \|\btheta\|_2 + \|\bhat{\btheta}\|_2 \right )}$, and as the ratio of misclassified entries vs.~total number of inliers. Both error metrics are tightly related, and all algorithms behave very similarly under both criteria.


In all our simulations, we use the ambient dimension $\d = 100$ and subspace dimension $\r = 5$. To represent the subspace $\sU$ and later on, to create a sample vector $\x$ in the subspace, we first generate a basis $\U \in \R{}^{\d \times \r}$ and a coefficient vector $\btheta \in \R{}^{\r}$ with i.i.d. standard gaussian entries.  Then we generate $\x$ using $\U\btheta + \epsilon$, where $\epsilon \in \R{}^{\d}$ refers to possible noise and is generated using i.i.d. ~$\mathscr{N}(0,\sigma^2)$. Finally, for each coordinate $\i \in [\d]$, with probability $p$, we replace the entry in $\x$ with an outlier generated using i.i.d standard gaussian distribution. For noisy formulation of MILP,  we solve \eqref{noiseEq} with $\lambdaa = 1000$. We use $T=50$ trials for all of our experiments. 

We implement the greedy algorithm in Matlab and for solving MILP formulation, we use CPLEX \cite{cplex} solver in AMPL environment and set a reasonable time limit (60 seconds) for each trial. CPLEX reports the best result found within the allowed amount of time. All of the experiments were run on a Macbook Pro with 3.5 GHz Intel Core i7 processor and 16 GB memory. 

\begin{figure} [!htbp]
\centering
 \resizebox{\columnwidth} {!} {\input{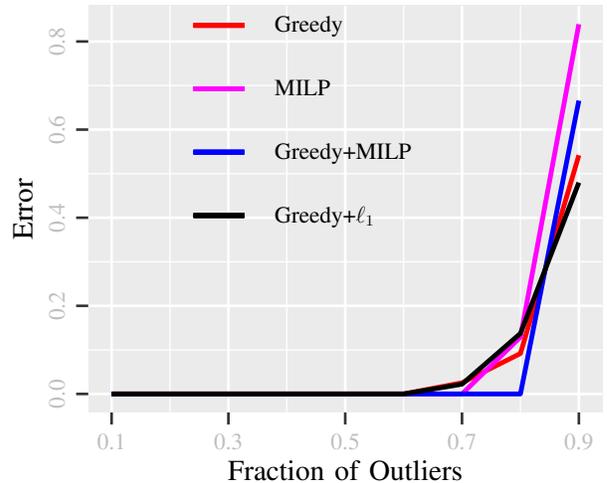}}
\caption{GLIMPS outperforms other approaches. In the first stage of GLIMPS, we remove 40\% outliers using the greedy algorithm and then in the second stage, we identify the remaining inliers and outliers using MILP. This approach succeeds for around 80\% outliers. In Greedy+$\ell_1$, we remove 100\% outliers (empirically fixed), then use $\ell_1$-minimization, but similar performance can be achieved using the greedy algorithm only.}
\label{FigGeneral}

\end{figure}

In our first experiment, we compare the performance of GLIMPS with other competing approaches as a function of the fraction of outliers. It is evident from Fig. \ref{FigGeneral} that GLIMPS outperforms all the other competing approaches. 

As we have shown in Fig. \ref{FigGeneral} that the combined performance of greedy and $\ell_1$-minimization can be achieved using only greedy algorithm that runs in $\Ord(\d^2)$, we only show comparative analysis of GLIMPS with greedy and MILP formulation in the subsequent experiments.

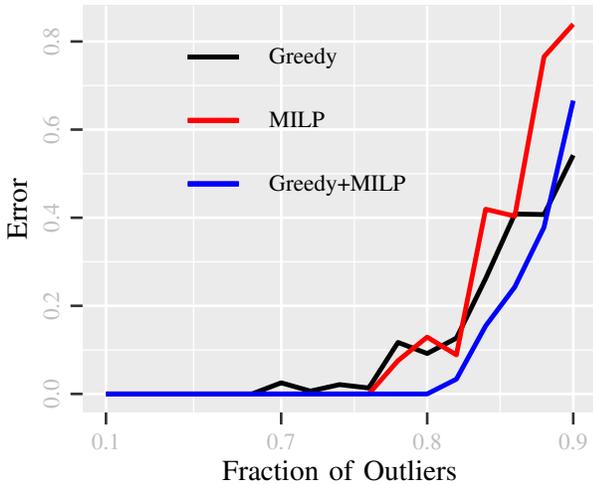
\begin{figure} [!htbp]
\centering
 \resizebox{\columnwidth} {!} {
\begin{tikzpicture}[x=1pt,y=1pt]
\definecolor{fillColor}{RGB}{255,255,255}
\path[use as bounding box,fill=fillColor,fill opacity=0.00] (0,0) rectangle (144.54,130.09);
\begin{scope}
\path[clip] (  0.00,  0.00) rectangle (144.54,130.09);
\definecolor{drawColor}{RGB}{255,255,255}
\definecolor{fillColor}{RGB}{255,255,255}

\path[draw=drawColor,line width= 0.6pt,line join=round,line cap=round,fill=fillColor] (  0.00, -0.00) rectangle (144.54,130.09);
\end{scope}
\begin{scope}
\path[clip] ( 21.94, 21.94) rectangle (139.04,114.67);
\definecolor{fillColor}{gray}{0.92}

\path[fill=fillColor] ( 21.94, 21.94) rectangle (139.04,114.67);
\definecolor{drawColor}{RGB}{255,255,255}

\path[draw=drawColor,line width= 0.3pt,line join=round] ( 21.94, 36.20) --
	(139.04, 36.20);

\path[draw=drawColor,line width= 0.3pt,line join=round] ( 21.94, 56.29) --
	(139.04, 56.29);

\path[draw=drawColor,line width= 0.3pt,line join=round] ( 21.94, 76.38) --
	(139.04, 76.38);

\path[draw=drawColor,line width= 0.3pt,line join=round] ( 21.94, 96.47) --
	(139.04, 96.47);

\path[draw=drawColor,line width= 0.3pt,line join=round] ( 47.23, 21.94) --
	( 47.23,114.67);

\path[draw=drawColor,line width= 0.3pt,line join=round] ( 83.82, 21.94) --
	( 83.82,114.67);

\path[draw=drawColor,line width= 0.3pt,line join=round] (117.08, 21.94) --
	(117.08,114.67);

\path[draw=drawColor,line width= 0.6pt,line join=round] ( 21.94, 26.16) --
	(139.04, 26.16);

\path[draw=drawColor,line width= 0.6pt,line join=round] ( 21.94, 46.25) --
	(139.04, 46.25);

\path[draw=drawColor,line width= 0.6pt,line join=round] ( 21.94, 66.34) --
	(139.04, 66.34);

\path[draw=drawColor,line width= 0.6pt,line join=round] ( 21.94, 86.43) --
	(139.04, 86.43);

\path[draw=drawColor,line width= 0.6pt,line join=round] ( 21.94,106.52) --
	(139.04,106.52);

\path[draw=drawColor,line width= 0.6pt,line join=round] ( 27.27, 21.94) --
	( 27.27,114.67);

\path[draw=drawColor,line width= 0.6pt,line join=round] ( 67.18, 21.94) --
	( 67.18,114.67);

\path[draw=drawColor,line width= 0.6pt,line join=round] (100.45, 21.94) --
	(100.45,114.67);

\path[draw=drawColor,line width= 0.6pt,line join=round] (133.72, 21.94) --
	(133.72,114.67);
\definecolor{drawColor}{RGB}{0,0,0}

\path[draw=drawColor,line width= 1.1pt,line join=round] ( 27.27, 26.16) --
	( 33.92, 26.16) --
	( 40.57, 26.16) --
	( 47.23, 26.16) --
	( 53.88, 26.16) --
	( 60.53, 26.16) --
	( 67.18, 28.70) --
	( 73.84, 26.81) --
	( 80.49, 28.29) --
	( 87.14, 27.52) --
	( 93.80, 37.91) --
	(100.45, 35.38) --
	(107.10, 38.88) --
	(113.76, 52.46) --
	(120.41, 67.19) --
	(127.06, 67.06) --
	(133.72, 80.59);
\definecolor{drawColor}{RGB}{255,0,0}

\path[draw=drawColor,line width= 1.1pt,line join=round] ( 27.27, 26.16) --
	( 33.92, 26.16) --
	( 40.57, 26.16) --
	( 47.23, 26.16) --
	( 53.88, 26.16) --
	( 60.53, 26.16) --
	( 67.18, 26.16) --
	( 73.84, 26.16) --
	( 80.49, 26.16) --
	( 87.14, 26.16) --
	( 93.80, 33.75) --
	(100.45, 39.12) --
	(107.10, 35.06) --
	(113.76, 68.28) --
	(120.41, 66.70) --
	(127.06,103.04) --
	(133.72,110.46);
\definecolor{drawColor}{RGB}{0,0,255}

\path[draw=drawColor,line width= 1.1pt,line join=round] ( 27.27, 26.16) --
	( 33.92, 26.16) --
	( 40.57, 26.16) --
	( 47.23, 26.16) --
	( 53.88, 26.16) --
	( 60.53, 26.16) --
	( 67.18, 26.16) --
	( 73.84, 26.16) --
	( 80.49, 26.16) --
	( 87.14, 26.16) --
	( 93.80, 26.16) --
	(100.45, 26.16) --
	(107.10, 29.53) --
	(113.76, 41.67) --
	(120.41, 50.56) --
	(127.06, 64.16) --
	(133.72, 93.04);
\end{scope}
\begin{scope}
\path[clip] (  0.00,  0.00) rectangle (144.54,130.09);
\definecolor{drawColor}{RGB}{190,190,190}

\node[text=drawColor,rotate= 90.00,anchor=base,inner sep=0pt, outer sep=0pt, scale=  0.44] at ( 16.57, 26.16) {0.0};

\node[text=drawColor,rotate= 90.00,anchor=base,inner sep=0pt, outer sep=0pt, scale=  0.44] at ( 16.57, 46.25) {0.2};

\node[text=drawColor,rotate= 90.00,anchor=base,inner sep=0pt, outer sep=0pt, scale=  0.44] at ( 16.57, 66.34) {0.4};

\node[text=drawColor,rotate= 90.00,anchor=base,inner sep=0pt, outer sep=0pt, scale=  0.44] at ( 16.57, 86.43) {0.6};

\node[text=drawColor,rotate= 90.00,anchor=base,inner sep=0pt, outer sep=0pt, scale=  0.44] at ( 16.57,106.52) {0.8};
\end{scope}
\begin{scope}
\path[clip] (  0.00,  0.00) rectangle (144.54,130.09);
\definecolor{drawColor}{gray}{0.20}

\path[draw=drawColor,line width= 0.6pt,line join=round] ( 19.19, 26.16) --
	( 21.94, 26.16);

\path[draw=drawColor,line width= 0.6pt,line join=round] ( 19.19, 46.25) --
	( 21.94, 46.25);

\path[draw=drawColor,line width= 0.6pt,line join=round] ( 19.19, 66.34) --
	( 21.94, 66.34);

\path[draw=drawColor,line width= 0.6pt,line join=round] ( 19.19, 86.43) --
	( 21.94, 86.43);

\path[draw=drawColor,line width= 0.6pt,line join=round] ( 19.19,106.52) --
	( 21.94,106.52);
\end{scope}
\begin{scope}
\path[clip] (  0.00,  0.00) rectangle (144.54,130.09);
\definecolor{drawColor}{gray}{0.20}

\path[draw=drawColor,line width= 0.6pt,line join=round] ( 27.27, 19.19) --
	( 27.27, 21.94);

\path[draw=drawColor,line width= 0.6pt,line join=round] ( 67.18, 19.19) --
	( 67.18, 21.94);

\path[draw=drawColor,line width= 0.6pt,line join=round] (100.45, 19.19) --
	(100.45, 21.94);

\path[draw=drawColor,line width= 0.6pt,line join=round] (133.72, 19.19) --
	(133.72, 21.94);
\end{scope}
\begin{scope}
\path[clip] (  0.00,  0.00) rectangle (144.54,130.09);
\definecolor{drawColor}{RGB}{190,190,190}

\node[text=drawColor,anchor=base,inner sep=0pt, outer sep=0pt, scale=  0.44] at ( 27.27, 13.53) {0.1};

\node[text=drawColor,anchor=base,inner sep=0pt, outer sep=0pt, scale=  0.44] at ( 67.18, 13.53) {0.7};

\node[text=drawColor,anchor=base,inner sep=0pt, outer sep=0pt, scale=  0.44] at (100.45, 13.53) {0.8};

\node[text=drawColor,anchor=base,inner sep=0pt, outer sep=0pt, scale=  0.44] at (133.72, 13.53) {0.9};
\end{scope}
\begin{scope}
\path[clip] (  0.00,  0.00) rectangle (144.54,130.09);
\definecolor{drawColor}{RGB}{0,0,0}

\node[text=drawColor,anchor=base,inner sep=0pt, outer sep=0pt, scale=  0.55] at ( 80.49,  6.57) {Fraction of Outliers};
\end{scope}
\begin{scope}
\path[clip] (  0.00,  0.00) rectangle (144.54,130.09);
\definecolor{drawColor}{RGB}{0,0,0}

\node[text=drawColor,rotate= 90.00,anchor=base,inner sep=0pt, outer sep=0pt, scale=  0.55] at (  9.29, 68.31) {Error};
\end{scope}
\begin{scope}
\path[clip] (  0.00,  0.00) rectangle (144.54,130.09);

\path[] ( 38.99, 61.34) rectangle ( 98.58,130.91);
\end{scope}
\begin{scope}
\path[clip] (  0.00,  0.00) rectangle (144.54,130.09);

\path[] ( 44.49, 95.74) rectangle ( 58.94,110.20);
\end{scope}
\begin{scope}
\path[clip] (  0.00,  0.00) rectangle (144.54,130.09);
\definecolor{drawColor}{RGB}{0,0,0}

\path[draw=drawColor,line width= 1.1pt,line join=round] ( 45.93,102.97) -- ( 57.50,102.97);
\end{scope}
\begin{scope}
\path[clip] (  0.00,  0.00) rectangle (144.54,130.09);
\definecolor{drawColor}{RGB}{0,0,0}

\path[draw=drawColor,line width= 1.1pt,line join=round] ( 45.93,102.97) -- ( 57.50,102.97);
\end{scope}
\begin{scope}
\path[clip] (  0.00,  0.00) rectangle (144.54,130.09);
\definecolor{drawColor}{RGB}{0,0,0}

\path[draw=drawColor,line width= 1.1pt,line join=round] ( 45.93,102.97) -- ( 57.50,102.97);
\end{scope}
\begin{scope}
\path[clip] (  0.00,  0.00) rectangle (144.54,130.09);

\path[] ( 44.49, 81.29) rectangle ( 58.94, 95.74);
\end{scope}
\begin{scope}
\path[clip] (  0.00,  0.00) rectangle (144.54,130.09);
\definecolor{drawColor}{RGB}{255,0,0}

\path[draw=drawColor,line width= 1.1pt,line join=round] ( 45.93, 88.52) -- ( 57.50, 88.52);
\end{scope}
\begin{scope}
\path[clip] (  0.00,  0.00) rectangle (144.54,130.09);
\definecolor{drawColor}{RGB}{255,0,0}

\path[draw=drawColor,line width= 1.1pt,line join=round] ( 45.93, 88.52) -- ( 57.50, 88.52);
\end{scope}
\begin{scope}
\path[clip] (  0.00,  0.00) rectangle (144.54,130.09);
\definecolor{drawColor}{RGB}{255,0,0}

\path[draw=drawColor,line width= 1.1pt,line join=round] ( 45.93, 88.52) -- ( 57.50, 88.52);
\end{scope}
\begin{scope}
\path[clip] (  0.00,  0.00) rectangle (144.54,130.09);

\path[] ( 44.49, 66.84) rectangle ( 58.94, 81.29);
\end{scope}
\begin{scope}
\path[clip] (  0.00,  0.00) rectangle (144.54,130.09);
\definecolor{drawColor}{RGB}{0,0,255}

\path[draw=drawColor,line width= 1.1pt,line join=round] ( 45.93, 74.06) -- ( 57.50, 74.06);
\end{scope}
\begin{scope}
\path[clip] (  0.00,  0.00) rectangle (144.54,130.09);
\definecolor{drawColor}{RGB}{0,0,255}

\path[draw=drawColor,line width= 1.1pt,line join=round] ( 45.93, 74.06) -- ( 57.50, 74.06);
\end{scope}
\begin{scope}
\path[clip] (  0.00,  0.00) rectangle (144.54,130.09);
\definecolor{drawColor}{RGB}{0,0,255}

\path[draw=drawColor,line width= 1.1pt,line join=round] ( 45.93, 74.06) -- ( 57.50, 74.06);
\end{scope}
\begin{scope}
\path[clip] (  0.00,  0.00) rectangle (144.54,130.09);
\definecolor{drawColor}{RGB}{0,0,0}

\node[text=drawColor,anchor=base west,inner sep=0pt, outer sep=0pt, scale=  0.44] at ( 64.44,101.46) {Greedy};
\end{scope}
\begin{scope}
\path[clip] (  0.00,  0.00) rectangle (144.54,130.09);
\definecolor{drawColor}{RGB}{0,0,0}

\node[text=drawColor,anchor=base west,inner sep=0pt, outer sep=0pt, scale=  0.44] at ( 64.44, 87.00) {MILP};
\end{scope}
\begin{scope}
\path[clip] (  0.00,  0.00) rectangle (144.54,130.09);
\definecolor{drawColor}{RGB}{0,0,0}

\node[text=drawColor,anchor=base west,inner sep=0pt, outer sep=0pt, scale=  0.44] at ( 64.44, 72.55) {Greedy+MILP};
\end{scope}
\begin{scope}
\path[clip] (  0.00,  0.00) rectangle (144.54,130.09);
\definecolor{drawColor}{RGB}{0,0,0}

\node[text=drawColor,anchor=base,inner sep=0pt, outer sep=0pt, scale=  0.50] at ( 80.49,121.14) {Performance of GLIMPS compared to Greedy and MILP};
\end{scope}
\end{tikzpicture}}
\caption{GLIMPS outperforms both the greedy and MILP approaches. In particular, GLIMPS tolerates 80\% outliers whereas MILP strictly fails after 76\%. Moreover, the performance of the greedy algorithm becomes unstable if the fraction of outliers goes beyond 60\%.}
\label{Fig40}
\end{figure}

In our second experiment, we show how the behavior of GLIMPS, the greedy {\em Erasure} algorithm and the MILP approach for \SRMSD\ change with a gradual increase in fraction of outliers. The average error rate for all algorithms is shown in Fig. \ref{Fig40}. It is observed that if we remove optimal number of outliers (in this case, 40\%) in the first stage, GLIMPS can achieve 100\% success rate for up to 80\% outliers, whereas the greedy algorithm and MILP approach can achieve similar performance (100\% success rate) for up to 60\% and 76\% outliers respectively.
 
\begin{figure}
\centering
 \resizebox{\columnwidth} {!} {
\begin{tikzpicture}[x=1pt,y=1pt]
\definecolor{fillColor}{RGB}{255,255,255}
\path[use as bounding box,fill=fillColor,fill opacity=0.00] (0,0) rectangle (144.54,130.09);
\begin{scope}
\path[clip] (  0.00,  0.00) rectangle (144.54,130.09);
\definecolor{drawColor}{RGB}{255,255,255}
\definecolor{fillColor}{RGB}{255,255,255}

\path[draw=drawColor,line width= 0.6pt,line join=round,line cap=round,fill=fillColor] (  0.00,  0.00) rectangle (144.54,130.09);
\end{scope}
\begin{scope}
\path[clip] ( 23.91, 23.91) rectangle (139.04,113.70);
\definecolor{fillColor}{gray}{0.92}

\path[fill=fillColor] ( 23.91, 23.91) rectangle (139.04,113.70);
\definecolor{drawColor}{RGB}{255,255,255}

\path[draw=drawColor,line width= 0.3pt,line join=round] ( 23.91, 40.96) --
	(139.04, 40.96);

\path[draw=drawColor,line width= 0.3pt,line join=round] ( 23.91, 67.55) --
	(139.04, 67.55);

\path[draw=drawColor,line width= 0.3pt,line join=round] ( 23.91, 94.14) --
	(139.04, 94.14);

\path[draw=drawColor,line width= 0.3pt,line join=round] ( 48.76, 23.91) --
	( 48.76,113.70);

\path[draw=drawColor,line width= 0.3pt,line join=round] ( 84.74, 23.91) --
	( 84.74,113.70);

\path[draw=drawColor,line width= 0.3pt,line join=round] (117.45, 23.91) --
	(117.45,113.70);

\path[draw=drawColor,line width= 0.6pt,line join=round] ( 23.91, 27.66) --
	(139.04, 27.66);

\path[draw=drawColor,line width= 0.6pt,line join=round] ( 23.91, 54.26) --
	(139.04, 54.26);

\path[draw=drawColor,line width= 0.6pt,line join=round] ( 23.91, 80.85) --
	(139.04, 80.85);

\path[draw=drawColor,line width= 0.6pt,line join=round] ( 23.91,107.44) --
	(139.04,107.44);

\path[draw=drawColor,line width= 0.6pt,line join=round] ( 29.14, 23.91) --
	( 29.14,113.70);

\path[draw=drawColor,line width= 0.6pt,line join=round] ( 68.39, 23.91) --
	( 68.39,113.70);

\path[draw=drawColor,line width= 0.6pt,line join=round] (101.10, 23.91) --
	(101.10,113.70);

\path[draw=drawColor,line width= 0.6pt,line join=round] (133.81, 23.91) --
	(133.81,113.70);
\definecolor{drawColor}{RGB}{0,0,0}

\path[draw=drawColor,line width= 1.1pt,line join=round] ( 29.14, 28.08) --
	( 35.68, 28.44) --
	( 42.22, 28.80) --
	( 48.76, 29.30) --
	( 55.31, 28.50) --
	( 61.85, 28.62) --
	( 68.39, 28.73) --
	( 74.93, 28.75) --
	( 81.47, 28.70) --
	( 88.01, 28.90) --
	( 94.56, 28.77) --
	(101.10, 28.77) --
	(107.64, 28.77) --
	(114.18, 28.84) --
	(120.72, 28.80) --
	(127.26, 28.91) --
	(133.81, 28.78);
\definecolor{drawColor}{RGB}{255,0,0}

\path[draw=drawColor,line width= 1.1pt,line join=round] ( 29.14, 28.12) --
	( 35.68, 64.86) --
	( 42.22,107.60) --
	( 48.76,107.71) --
	( 55.31,107.84) --
	( 61.85,108.06) --
	( 68.39,108.11) --
	( 74.93,107.93) --
	( 81.47,108.02) --
	( 88.01,107.98) --
	( 94.56,107.92) --
	(101.10,107.98) --
	(107.64,108.03) --
	(114.18,108.02) --
	(120.72,108.04) --
	(127.26,107.95) --
	(133.81,107.96);
\definecolor{drawColor}{RGB}{0,0,255}

\path[draw=drawColor,line width= 1.1pt,line join=round] ( 29.14, 27.99) --
	( 35.68, 29.06) --
	( 42.22, 46.19) --
	( 48.76,101.73) --
	( 55.31,105.94) --
	( 61.85,108.43) --
	( 68.39,108.85) --
	( 74.93,109.02) --
	( 81.47,108.93) --
	( 88.01,109.12) --
	( 94.56,109.17) --
	(101.10,109.22) --
	(107.64,109.23) --
	(114.18,109.36) --
	(120.72,109.45) --
	(127.26,109.56) --
	(133.81,109.62);
\end{scope}
\begin{scope}
\path[clip] (  0.00,  0.00) rectangle (144.54,130.09);
\definecolor{drawColor}{RGB}{190,190,190}

\node[text=drawColor,rotate= 90.00,anchor=base,inner sep=0pt, outer sep=0pt, scale=  0.44] at ( 17.98, 27.66) {0};

\node[text=drawColor,rotate= 90.00,anchor=base,inner sep=0pt, outer sep=0pt, scale=  0.44] at ( 17.98, 54.26) {20};

\node[text=drawColor,rotate= 90.00,anchor=base,inner sep=0pt, outer sep=0pt, scale=  0.44] at ( 17.98, 80.85) {40};

\node[text=drawColor,rotate= 90.00,anchor=base,inner sep=0pt, outer sep=0pt, scale=  0.44] at ( 17.98,107.44) {60};
\end{scope}
\begin{scope}
\path[clip] (  0.00,  0.00) rectangle (144.54,130.09);
\definecolor{drawColor}{gray}{0.20}

\path[draw=drawColor,line width= 0.6pt,line join=round] ( 21.16, 27.66) --
	( 23.91, 27.66);

\path[draw=drawColor,line width= 0.6pt,line join=round] ( 21.16, 54.26) --
	( 23.91, 54.26);

\path[draw=drawColor,line width= 0.6pt,line join=round] ( 21.16, 80.85) --
	( 23.91, 80.85);

\path[draw=drawColor,line width= 0.6pt,line join=round] ( 21.16,107.44) --
	( 23.91,107.44);
\end{scope}
\begin{scope}
\path[clip] (  0.00,  0.00) rectangle (144.54,130.09);
\definecolor{drawColor}{gray}{0.20}

\path[draw=drawColor,line width= 0.6pt,line join=round] ( 29.14, 21.16) --
	( 29.14, 23.91);

\path[draw=drawColor,line width= 0.6pt,line join=round] ( 68.39, 21.16) --
	( 68.39, 23.91);

\path[draw=drawColor,line width= 0.6pt,line join=round] (101.10, 21.16) --
	(101.10, 23.91);

\path[draw=drawColor,line width= 0.6pt,line join=round] (133.81, 21.16) --
	(133.81, 23.91);
\end{scope}
\begin{scope}
\path[clip] (  0.00,  0.00) rectangle (144.54,130.09);
\definecolor{drawColor}{RGB}{190,190,190}

\node[text=drawColor,anchor=base,inner sep=0pt, outer sep=0pt, scale=  0.44] at ( 29.14, 14.95) {0.1};

\node[text=drawColor,anchor=base,inner sep=0pt, outer sep=0pt, scale=  0.44] at ( 68.39, 14.95) {0.7};

\node[text=drawColor,anchor=base,inner sep=0pt, outer sep=0pt, scale=  0.44] at (101.10, 14.95) {0.8};

\node[text=drawColor,anchor=base,inner sep=0pt, outer sep=0pt, scale=  0.44] at (133.81, 14.95) {0.9};
\end{scope}
\begin{scope}
\path[clip] (  0.00,  0.00) rectangle (144.54,130.09);
\definecolor{drawColor}{RGB}{0,0,0}

\node[text=drawColor,anchor=base,inner sep=0pt, outer sep=0pt, scale=  0.55] at ( 81.47,  7.44) {Fraction of Outliers};
\end{scope}
\begin{scope}
\path[clip] (  0.00,  0.00) rectangle (144.54,130.09);
\definecolor{drawColor}{RGB}{0,0,0}

\node[text=drawColor,rotate= 90.00,anchor=base,inner sep=0pt, outer sep=0pt, scale=  0.55] at (  9.29, 68.80) {Computation Time (seconds)};
\end{scope}
\begin{scope}
\path[clip] (  0.00,  0.00) rectangle (144.54,130.09);

\path[] ( 86.22, 25.13) rectangle (145.81, 94.51);
\end{scope}
\begin{scope}
\path[clip] (  0.00,  0.00) rectangle (144.54,130.09);

\path[] ( 91.72, 59.54) rectangle (106.17, 73.99);
\end{scope}
\begin{scope}
\path[clip] (  0.00,  0.00) rectangle (144.54,130.09);
\definecolor{drawColor}{RGB}{0,0,0}

\path[draw=drawColor,line width= 1.1pt,line join=round] ( 93.16, 66.77) -- (104.73, 66.77);
\end{scope}
\begin{scope}
\path[clip] (  0.00,  0.00) rectangle (144.54,130.09);
\definecolor{drawColor}{RGB}{0,0,0}

\path[draw=drawColor,line width= 1.1pt,line join=round] ( 93.16, 66.77) -- (104.73, 66.77);
\end{scope}
\begin{scope}
\path[clip] (  0.00,  0.00) rectangle (144.54,130.09);
\definecolor{drawColor}{RGB}{0,0,0}

\path[draw=drawColor,line width= 1.1pt,line join=round] ( 93.16, 66.77) -- (104.73, 66.77);
\end{scope}
\begin{scope}
\path[clip] (  0.00,  0.00) rectangle (144.54,130.09);

\path[] ( 91.72, 45.09) rectangle (106.17, 59.54);
\end{scope}
\begin{scope}
\path[clip] (  0.00,  0.00) rectangle (144.54,130.09);
\definecolor{drawColor}{RGB}{255,0,0}

\path[draw=drawColor,line width= 1.1pt,line join=round] ( 93.16, 52.31) -- (104.73, 52.31);
\end{scope}
\begin{scope}
\path[clip] (  0.00,  0.00) rectangle (144.54,130.09);
\definecolor{drawColor}{RGB}{255,0,0}

\path[draw=drawColor,line width= 1.1pt,line join=round] ( 93.16, 52.31) -- (104.73, 52.31);
\end{scope}
\begin{scope}
\path[clip] (  0.00,  0.00) rectangle (144.54,130.09);
\definecolor{drawColor}{RGB}{255,0,0}

\path[draw=drawColor,line width= 1.1pt,line join=round] ( 93.16, 52.31) -- (104.73, 52.31);
\end{scope}
\begin{scope}
\path[clip] (  0.00,  0.00) rectangle (144.54,130.09);

\path[] ( 91.72, 30.63) rectangle (106.17, 45.09);
\end{scope}
\begin{scope}
\path[clip] (  0.00,  0.00) rectangle (144.54,130.09);
\definecolor{drawColor}{RGB}{0,0,255}

\path[draw=drawColor,line width= 1.1pt,line join=round] ( 93.16, 37.86) -- (104.73, 37.86);
\end{scope}
\begin{scope}
\path[clip] (  0.00,  0.00) rectangle (144.54,130.09);
\definecolor{drawColor}{RGB}{0,0,255}

\path[draw=drawColor,line width= 1.1pt,line join=round] ( 93.16, 37.86) -- (104.73, 37.86);
\end{scope}
\begin{scope}
\path[clip] (  0.00,  0.00) rectangle (144.54,130.09);
\definecolor{drawColor}{RGB}{0,0,255}

\path[draw=drawColor,line width= 1.1pt,line join=round] ( 93.16, 37.86) -- (104.73, 37.86);
\end{scope}
\begin{scope}
\path[clip] (  0.00,  0.00) rectangle (144.54,130.09);
\definecolor{drawColor}{RGB}{0,0,0}

\node[text=drawColor,anchor=base west,inner sep=0pt, outer sep=0pt, scale=  0.44] at (111.67, 65.25) {Greedy};
\end{scope}
\begin{scope}
\path[clip] (  0.00,  0.00) rectangle (144.54,130.09);
\definecolor{drawColor}{RGB}{0,0,0}

\node[text=drawColor,anchor=base west,inner sep=0pt, outer sep=0pt, scale=  0.44] at (111.67, 50.80) {MILP};
\end{scope}
\begin{scope}
\path[clip] (  0.00,  0.00) rectangle (144.54,130.09);
\definecolor{drawColor}{RGB}{0,0,0}

\node[text=drawColor,anchor=base west,inner sep=0pt, outer sep=0pt, scale=  0.44] at (111.67, 36.34) {Greedy+MILP};
\end{scope}
\begin{scope}
\path[clip] (  0.00,  0.00) rectangle (144.54,130.09);
\definecolor{drawColor}{RGB}{0,0,0}

\node[text=drawColor,anchor=base,inner sep=0pt, outer sep=0pt, scale=  0.50] at ( 81.47,121.14) {Analysis of computation time requirements };
\end{scope}
\end{tikzpicture}}
\caption{MILP and GLIMPS take longer time to solve the problem. The greedy algorithm uses heuristics to reduce the solution space and hence computationally inexpensive. In this experiment, we remove 40\% in the first stage of GLIMPS.}
\label{FigT}
\end{figure}
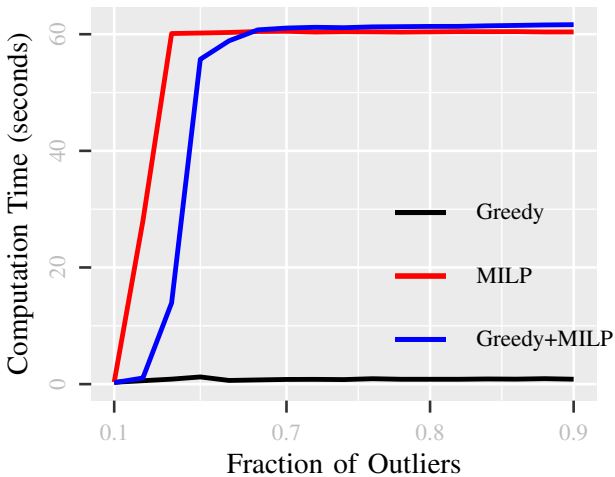

In our third experiment, we study the computational time requirements for the greedy, MILP and proposed GLIMPS approaches. As this problem is a combinatorial problem, MILP and GLIMPS take longer time when compared with the greedy algorithm, which reduces the search space using heuristics and thus takes significantly less time (polynomial time). The results are shown in Fig. \ref{FigT}.

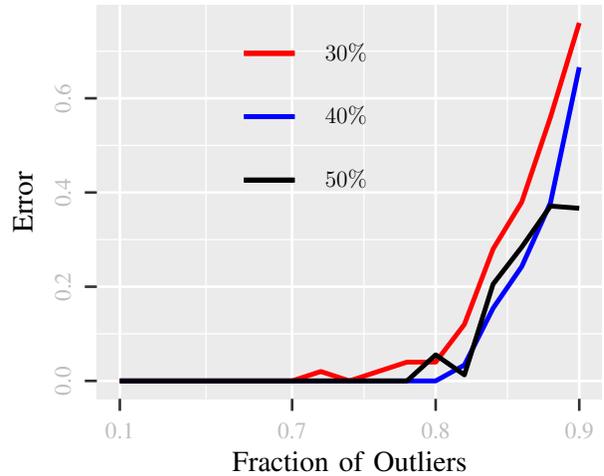
\begin{figure}
\centering
 \resizebox{\columnwidth} {!} {
\begin{tikzpicture}[x=1pt,y=1pt]
\definecolor{fillColor}{RGB}{255,255,255}
\path[use as bounding box,fill=fillColor,fill opacity=0.00] (0,0) rectangle (144.54,130.09);
\begin{scope}
\path[clip] (  0.00,  0.00) rectangle (144.54,130.09);
\definecolor{drawColor}{RGB}{255,255,255}
\definecolor{fillColor}{RGB}{255,255,255}

\path[draw=drawColor,line width= 0.6pt,line join=round,line cap=round,fill=fillColor] (  0.00,  0.00) rectangle (144.54,130.09);
\end{scope}
\begin{scope}
\path[clip] ( 23.91, 23.91) rectangle (139.04,113.70);
\definecolor{fillColor}{gray}{0.92}

\path[fill=fillColor] ( 23.91, 23.91) rectangle (139.04,113.70);
\definecolor{drawColor}{RGB}{255,255,255}

\path[draw=drawColor,line width= 0.3pt,line join=round] ( 23.91, 38.73) --
	(139.04, 38.73);

\path[draw=drawColor,line width= 0.3pt,line join=round] ( 23.91, 60.21) --
	(139.04, 60.21);

\path[draw=drawColor,line width= 0.3pt,line join=round] ( 23.91, 81.69) --
	(139.04, 81.69);

\path[draw=drawColor,line width= 0.3pt,line join=round] ( 23.91,103.17) --
	(139.04,103.17);

\path[draw=drawColor,line width= 0.3pt,line join=round] ( 48.76, 23.91) --
	( 48.76,113.70);

\path[draw=drawColor,line width= 0.3pt,line join=round] ( 84.74, 23.91) --
	( 84.74,113.70);

\path[draw=drawColor,line width= 0.3pt,line join=round] (117.45, 23.91) --
	(117.45,113.70);

\path[draw=drawColor,line width= 0.6pt,line join=round] ( 23.91, 27.99) --
	(139.04, 27.99);

\path[draw=drawColor,line width= 0.6pt,line join=round] ( 23.91, 49.47) --
	(139.04, 49.47);

\path[draw=drawColor,line width= 0.6pt,line join=round] ( 23.91, 70.95) --
	(139.04, 70.95);

\path[draw=drawColor,line width= 0.6pt,line join=round] ( 23.91, 92.43) --
	(139.04, 92.43);

\path[draw=drawColor,line width= 0.6pt,line join=round] ( 29.14, 23.91) --
	( 29.14,113.70);

\path[draw=drawColor,line width= 0.6pt,line join=round] ( 68.39, 23.91) --
	( 68.39,113.70);

\path[draw=drawColor,line width= 0.6pt,line join=round] (101.10, 23.91) --
	(101.10,113.70);

\path[draw=drawColor,line width= 0.6pt,line join=round] (133.81, 23.91) --
	(133.81,113.70);
\definecolor{drawColor}{RGB}{255,0,0}

\path[draw=drawColor,line width= 1.1pt,line join=round] ( 29.14, 27.99) --
	( 35.68, 27.99) --
	( 42.22, 27.99) --
	( 48.76, 27.99) --
	( 55.31, 27.99) --
	( 61.85, 27.99) --
	( 68.39, 27.99) --
	( 74.93, 30.14) --
	( 81.47, 27.99) --
	( 88.01, 30.14) --
	( 94.56, 32.28) --
	(101.10, 32.28) --
	(107.64, 40.88) --
	(114.18, 58.06) --
	(120.72, 68.80) --
	(127.26, 88.14) --
	(133.81,109.62);
\definecolor{drawColor}{RGB}{0,0,255}

\path[draw=drawColor,line width= 1.1pt,line join=round] ( 29.14, 27.99) --
	( 35.68, 27.99) --
	( 42.22, 27.99) --
	( 48.76, 27.99) --
	( 55.31, 27.99) --
	( 61.85, 27.99) --
	( 68.39, 27.99) --
	( 74.93, 27.99) --
	( 81.47, 27.99) --
	( 88.01, 27.99) --
	( 94.56, 27.99) --
	(101.10, 27.99) --
	(107.64, 31.60) --
	(114.18, 44.57) --
	(120.72, 54.08) --
	(127.26, 68.62) --
	(133.81, 99.50);
\definecolor{drawColor}{RGB}{0,0,0}

\path[draw=drawColor,line width= 1.1pt,line join=round] ( 29.14, 27.99) --
	( 35.68, 27.99) --
	( 42.22, 27.99) --
	( 48.76, 27.99) --
	( 55.31, 27.99) --
	( 61.85, 27.99) --
	( 68.39, 27.99) --
	( 74.93, 27.99) --
	( 81.47, 27.99) --
	( 88.01, 27.99) --
	( 94.56, 27.99) --
	(101.10, 33.94) --
	(107.64, 29.36) --
	(114.18, 50.06) --
	(120.72, 58.52) --
	(127.26, 67.86) --
	(133.81, 67.35);
\end{scope}
\begin{scope}
\path[clip] (  0.00,  0.00) rectangle (144.54,130.09);
\definecolor{drawColor}{RGB}{190,190,190}

\node[text=drawColor,rotate= 90.00,anchor=base,inner sep=0pt, outer sep=0pt, scale=  0.44] at ( 17.98, 27.99) {0.0};

\node[text=drawColor,rotate= 90.00,anchor=base,inner sep=0pt, outer sep=0pt, scale=  0.44] at ( 17.98, 49.47) {0.2};

\node[text=drawColor,rotate= 90.00,anchor=base,inner sep=0pt, outer sep=0pt, scale=  0.44] at ( 17.98, 70.95) {0.4};

\node[text=drawColor,rotate= 90.00,anchor=base,inner sep=0pt, outer sep=0pt, scale=  0.44] at ( 17.98, 92.43) {0.6};
\end{scope}
\begin{scope}
\path[clip] (  0.00,  0.00) rectangle (144.54,130.09);
\definecolor{drawColor}{gray}{0.20}

\path[draw=drawColor,line width= 0.6pt,line join=round] ( 21.16, 27.99) --
	( 23.91, 27.99);

\path[draw=drawColor,line width= 0.6pt,line join=round] ( 21.16, 49.47) --
	( 23.91, 49.47);

\path[draw=drawColor,line width= 0.6pt,line join=round] ( 21.16, 70.95) --
	( 23.91, 70.95);

\path[draw=drawColor,line width= 0.6pt,line join=round] ( 21.16, 92.43) --
	( 23.91, 92.43);
\end{scope}
\begin{scope}
\path[clip] (  0.00,  0.00) rectangle (144.54,130.09);
\definecolor{drawColor}{gray}{0.20}

\path[draw=drawColor,line width= 0.6pt,line join=round] ( 29.14, 21.16) --
	( 29.14, 23.91);

\path[draw=drawColor,line width= 0.6pt,line join=round] ( 68.39, 21.16) --
	( 68.39, 23.91);

\path[draw=drawColor,line width= 0.6pt,line join=round] (101.10, 21.16) --
	(101.10, 23.91);

\path[draw=drawColor,line width= 0.6pt,line join=round] (133.81, 21.16) --
	(133.81, 23.91);
\end{scope}
\begin{scope}
\path[clip] (  0.00,  0.00) rectangle (144.54,130.09);
\definecolor{drawColor}{RGB}{190,190,190}

\node[text=drawColor,anchor=base,inner sep=0pt, outer sep=0pt, scale=  0.44] at ( 29.14, 14.95) {0.1};

\node[text=drawColor,anchor=base,inner sep=0pt, outer sep=0pt, scale=  0.44] at ( 68.39, 14.95) {0.7};

\node[text=drawColor,anchor=base,inner sep=0pt, outer sep=0pt, scale=  0.44] at (101.10, 14.95) {0.8};

\node[text=drawColor,anchor=base,inner sep=0pt, outer sep=0pt, scale=  0.44] at (133.81, 14.95) {0.9};
\end{scope}
\begin{scope}
\path[clip] (  0.00,  0.00) rectangle (144.54,130.09);
\definecolor{drawColor}{RGB}{0,0,0}

\node[text=drawColor,anchor=base,inner sep=0pt, outer sep=0pt, scale=  0.55] at ( 81.47,  7.44) {Fraction of Outliers};
\end{scope}
\begin{scope}
\path[clip] (  0.00,  0.00) rectangle (144.54,130.09);
\definecolor{drawColor}{RGB}{0,0,0}

\node[text=drawColor,rotate= 90.00,anchor=base,inner sep=0pt, outer sep=0pt, scale=  0.55] at (  9.29, 68.80) {Error};
\end{scope}
\begin{scope}
\path[clip] (  0.00,  0.00) rectangle (144.54,130.09);

\path[] ( 50.45, 61.05) rectangle ( 89.47,130.43);
\end{scope}
\begin{scope}
\path[clip] (  0.00,  0.00) rectangle (144.54,130.09);

\path[] ( 55.95, 95.46) rectangle ( 70.40,109.91);
\end{scope}
\begin{scope}
\path[clip] (  0.00,  0.00) rectangle (144.54,130.09);
\definecolor{drawColor}{RGB}{255,0,0}

\path[draw=drawColor,line width= 1.1pt,line join=round] ( 57.39,102.68) -- ( 68.96,102.68);
\end{scope}
\begin{scope}
\path[clip] (  0.00,  0.00) rectangle (144.54,130.09);
\definecolor{drawColor}{RGB}{255,0,0}

\path[draw=drawColor,line width= 1.1pt,line join=round] ( 57.39,102.68) -- ( 68.96,102.68);
\end{scope}
\begin{scope}
\path[clip] (  0.00,  0.00) rectangle (144.54,130.09);
\definecolor{drawColor}{RGB}{255,0,0}

\path[draw=drawColor,line width= 1.1pt,line join=round] ( 57.39,102.68) -- ( 68.96,102.68);
\end{scope}
\begin{scope}
\path[clip] (  0.00,  0.00) rectangle (144.54,130.09);

\path[] ( 55.95, 81.00) rectangle ( 70.40, 95.46);
\end{scope}
\begin{scope}
\path[clip] (  0.00,  0.00) rectangle (144.54,130.09);
\definecolor{drawColor}{RGB}{0,0,255}

\path[draw=drawColor,line width= 1.1pt,line join=round] ( 57.39, 88.23) -- ( 68.96, 88.23);
\end{scope}
\begin{scope}
\path[clip] (  0.00,  0.00) rectangle (144.54,130.09);
\definecolor{drawColor}{RGB}{0,0,255}

\path[draw=drawColor,line width= 1.1pt,line join=round] ( 57.39, 88.23) -- ( 68.96, 88.23);
\end{scope}
\begin{scope}
\path[clip] (  0.00,  0.00) rectangle (144.54,130.09);
\definecolor{drawColor}{RGB}{0,0,255}

\path[draw=drawColor,line width= 1.1pt,line join=round] ( 57.39, 88.23) -- ( 68.96, 88.23);
\end{scope}
\begin{scope}
\path[clip] (  0.00,  0.00) rectangle (144.54,130.09);

\path[] ( 55.95, 66.55) rectangle ( 70.40, 81.00);
\end{scope}
\begin{scope}
\path[clip] (  0.00,  0.00) rectangle (144.54,130.09);
\definecolor{drawColor}{RGB}{0,0,0}

\path[draw=drawColor,line width= 1.1pt,line join=round] ( 57.39, 73.78) -- ( 68.96, 73.78);
\end{scope}
\begin{scope}
\path[clip] (  0.00,  0.00) rectangle (144.54,130.09);
\definecolor{drawColor}{RGB}{0,0,0}

\path[draw=drawColor,line width= 1.1pt,line join=round] ( 57.39, 73.78) -- ( 68.96, 73.78);
\end{scope}
\begin{scope}
\path[clip] (  0.00,  0.00) rectangle (144.54,130.09);
\definecolor{drawColor}{RGB}{0,0,0}

\path[draw=drawColor,line width= 1.1pt,line join=round] ( 57.39, 73.78) -- ( 68.96, 73.78);
\end{scope}
\begin{scope}
\path[clip] (  0.00,  0.00) rectangle (144.54,130.09);
\definecolor{drawColor}{RGB}{0,0,0}

\node[text=drawColor,anchor=base west,inner sep=0pt, outer sep=0pt, scale=  0.44] at ( 75.90,101.17) {$30\%$};
\end{scope}
\begin{scope}
\path[clip] (  0.00,  0.00) rectangle (144.54,130.09);
\definecolor{drawColor}{RGB}{0,0,0}

\node[text=drawColor,anchor=base west,inner sep=0pt, outer sep=0pt, scale=  0.44] at ( 75.90, 86.71) {$40\%$};
\end{scope}
\begin{scope}
\path[clip] (  0.00,  0.00) rectangle (144.54,130.09);
\definecolor{drawColor}{RGB}{0,0,0}

\node[text=drawColor,anchor=base west,inner sep=0pt, outer sep=0pt, scale=  0.44] at ( 75.90, 72.26) {$50\%$};
\end{scope}
\begin{scope}
\path[clip] (  0.00,  0.00) rectangle (144.54,130.09);
\definecolor{drawColor}{RGB}{0,0,0}

\node[text=drawColor,anchor=base,inner sep=0pt, outer sep=0pt, scale=  0.50] at ( 81.47,121.14) {Effect of removal percentage in greedy stage};
\end{scope}
\end{tikzpicture}}
\caption{We show that 40\% removal in the first stage provides the best result in the second stage. Removal of 30\% is low, whereas removal of 50\% is ambitious in abundant-outliers regime. All three removal quantities perform very well in sparse-outliers regime.}
\label{remRate}
\end{figure}

In our fourth experiment, we study the significance of outlier removal percentage in the first stage (greedy stage) of GLIMPS. We use this experiment as a step of hyperparameter tuning and apply three different removal percentages, namely 30\%, 40\% and 50\%. The average results are shown in Fig. \ref{remRate}.  It is noticed that 40\% removal of outliers in the greedy stage achieves the best tolerance of outliers. 

\begin{figure*}
\centering
\hspace{-2cm}
\begin{tikzpicture}[x=1pt,y=1pt]
\definecolor{fillColor}{RGB}{255,255,255}
\path[use as bounding box,fill=fillColor,fill opacity=0.00] (0,0) rectangle (144.54,130.09);
\begin{scope}
\path[clip] (  0.00,  0.00) rectangle (144.54,130.09);
\definecolor{drawColor}{RGB}{255,255,255}
\definecolor{fillColor}{RGB}{255,255,255}

\path[draw=drawColor,line width= 0.6pt,line join=round,line cap=round,fill=fillColor] (  0.00,  0.00) rectangle (144.54,130.09);
\end{scope}
\begin{scope}
\path[clip] ( 28.15, 29.36) rectangle (139.04,110.25);
\definecolor{fillColor}{gray}{0.92}

\path[fill=fillColor] ( 28.15, 29.36) rectangle (139.04,110.25);
\definecolor{drawColor}{RGB}{255,255,255}

\path[draw=drawColor,line width= 0.3pt,line join=round] ( 28.15, 43.71) --
	(139.04, 43.71);

\path[draw=drawColor,line width= 0.3pt,line join=round] ( 28.15, 65.09) --
	(139.04, 65.09);

\path[draw=drawColor,line width= 0.3pt,line join=round] ( 28.15, 86.48) --
	(139.04, 86.48);

\path[draw=drawColor,line width= 0.3pt,line join=round] ( 28.15,107.86) --
	(139.04,107.86);

\path[draw=drawColor,line width= 0.3pt,line join=round] ( 39.49, 29.36) --
	( 39.49,110.25);

\path[draw=drawColor,line width= 0.3pt,line join=round] ( 52.09, 29.36) --
	( 52.09,110.25);

\path[draw=drawColor,line width= 0.3pt,line join=round] ( 64.69, 29.36) --
	( 64.69,110.25);

\path[draw=drawColor,line width= 0.3pt,line join=round] ( 77.29, 29.36) --
	( 77.29,110.25);

\path[draw=drawColor,line width= 0.3pt,line join=round] ( 89.90, 29.36) --
	( 89.90,110.25);

\path[draw=drawColor,line width= 0.3pt,line join=round] (102.50, 29.36) --
	(102.50,110.25);

\path[draw=drawColor,line width= 0.3pt,line join=round] (115.10, 29.36) --
	(115.10,110.25);

\path[draw=drawColor,line width= 0.3pt,line join=round] (127.70, 29.36) --
	(127.70,110.25);

\path[draw=drawColor,line width= 0.6pt,line join=round] ( 28.15, 33.02) --
	(139.04, 33.02);

\path[draw=drawColor,line width= 0.6pt,line join=round] ( 28.15, 54.40) --
	(139.04, 54.40);

\path[draw=drawColor,line width= 0.6pt,line join=round] ( 28.15, 75.78) --
	(139.04, 75.78);

\path[draw=drawColor,line width= 0.6pt,line join=round] ( 28.15, 97.17) --
	(139.04, 97.17);

\path[draw=drawColor,line width= 0.6pt,line join=round] ( 33.19, 29.36) --
	( 33.19,110.25);

\path[draw=drawColor,line width= 0.6pt,line join=round] ( 45.79, 29.36) --
	( 45.79,110.25);

\path[draw=drawColor,line width= 0.6pt,line join=round] ( 58.39, 29.36) --
	( 58.39,110.25);

\path[draw=drawColor,line width= 0.6pt,line join=round] ( 70.99, 29.36) --
	( 70.99,110.25);

\path[draw=drawColor,line width= 0.6pt,line join=round] ( 83.59, 29.36) --
	( 83.59,110.25);

\path[draw=drawColor,line width= 0.6pt,line join=round] ( 96.20, 29.36) --
	( 96.20,110.25);

\path[draw=drawColor,line width= 0.6pt,line join=round] (108.80, 29.36) --
	(108.80,110.25);

\path[draw=drawColor,line width= 0.6pt,line join=round] (121.40, 29.36) --
	(121.40,110.25);

\path[draw=drawColor,line width= 0.6pt,line join=round] (134.00, 29.36) --
	(134.00,110.25);
\definecolor{drawColor}{RGB}{0,0,0}

\path[draw=drawColor,line width= 1.7pt,line join=round] ( 33.19, 33.06) --
	( 45.79, 33.19) --
	( 58.39, 33.24) --
	( 70.99, 33.33) --
	( 83.59, 33.41) --
	( 96.20, 33.66) --
	(108.80, 34.44) --
	(121.40, 41.78) --
	(134.00,106.58);
\definecolor{drawColor}{RGB}{0,0,255}

\path[draw=drawColor,line width= 1.7pt,line join=round] ( 33.19, 33.05) --
	( 45.79, 33.04) --
	( 58.39, 33.09) --
	( 70.99, 33.20) --
	( 83.59, 33.22) --
	( 96.20, 33.29) --
	(108.80, 33.47) --
	(121.40, 37.17) --
	(134.00, 88.62);
\end{scope}
\begin{scope}
\path[clip] (  0.00,  0.00) rectangle (144.54,130.09);
\definecolor{drawColor}{RGB}{190,190,190}

\node[text=drawColor,rotate= 90.00,anchor=base,inner sep=0pt, outer sep=0pt, scale=  0.62] at ( 22.23, 33.02) {0.0};

\node[text=drawColor,rotate= 90.00,anchor=base,inner sep=0pt, outer sep=0pt, scale=  0.62] at ( 22.23, 54.40) {0.5};

\node[text=drawColor,rotate= 90.00,anchor=base,inner sep=0pt, outer sep=0pt, scale=  0.62] at ( 22.23, 75.78) {1.0};

\node[text=drawColor,rotate= 90.00,anchor=base,inner sep=0pt, outer sep=0pt, scale=  0.62] at ( 22.23, 97.17) {1.5};
\end{scope}
\begin{scope}
\path[clip] (  0.00,  0.00) rectangle (144.54,130.09);
\definecolor{drawColor}{gray}{0.20}

\path[draw=drawColor,line width= 0.6pt,line join=round] ( 25.40, 33.02) --
	( 28.15, 33.02);

\path[draw=drawColor,line width= 0.6pt,line join=round] ( 25.40, 54.40) --
	( 28.15, 54.40);

\path[draw=drawColor,line width= 0.6pt,line join=round] ( 25.40, 75.78) --
	( 28.15, 75.78);

\path[draw=drawColor,line width= 0.6pt,line join=round] ( 25.40, 97.17) --
	( 28.15, 97.17);
\end{scope}
\begin{scope}
\path[clip] (  0.00,  0.00) rectangle (144.54,130.09);
\definecolor{drawColor}{gray}{0.20}

\path[draw=drawColor,line width= 0.6pt,line join=round] ( 33.19, 26.61) --
	( 33.19, 29.36);

\path[draw=drawColor,line width= 0.6pt,line join=round] ( 45.79, 26.61) --
	( 45.79, 29.36);

\path[draw=drawColor,line width= 0.6pt,line join=round] ( 58.39, 26.61) --
	( 58.39, 29.36);

\path[draw=drawColor,line width= 0.6pt,line join=round] ( 70.99, 26.61) --
	( 70.99, 29.36);

\path[draw=drawColor,line width= 0.6pt,line join=round] ( 83.59, 26.61) --
	( 83.59, 29.36);

\path[draw=drawColor,line width= 0.6pt,line join=round] ( 96.20, 26.61) --
	( 96.20, 29.36);

\path[draw=drawColor,line width= 0.6pt,line join=round] (108.80, 26.61) --
	(108.80, 29.36);

\path[draw=drawColor,line width= 0.6pt,line join=round] (121.40, 26.61) --
	(121.40, 29.36);

\path[draw=drawColor,line width= 0.6pt,line join=round] (134.00, 26.61) --
	(134.00, 29.36);
\end{scope}
\begin{scope}
\path[clip] (  0.00,  0.00) rectangle (144.54,130.09);
\definecolor{drawColor}{RGB}{190,190,190}

\node[text=drawColor,anchor=base,inner sep=0pt, outer sep=0pt, scale=  0.79] at ( 33.19, 17.98) {0.1};

\node[text=drawColor,anchor=base,inner sep=0pt, outer sep=0pt, scale=  0.79] at ( 45.79, 17.98) {0.2};

\node[text=drawColor,anchor=base,inner sep=0pt, outer sep=0pt, scale=  0.79] at ( 58.39, 17.98) {0.3};

\node[text=drawColor,anchor=base,inner sep=0pt, outer sep=0pt, scale=  0.79] at ( 70.99, 17.98) {0.4};

\node[text=drawColor,anchor=base,inner sep=0pt, outer sep=0pt, scale=  0.79] at ( 83.59, 17.98) {0.5};

\node[text=drawColor,anchor=base,inner sep=0pt, outer sep=0pt, scale=  0.79] at ( 96.20, 17.98) {0.6};

\node[text=drawColor,anchor=base,inner sep=0pt, outer sep=0pt, scale=  0.79] at (108.80, 17.98) {0.7};

\node[text=drawColor,anchor=base,inner sep=0pt, outer sep=0pt, scale=  0.79] at (121.40, 17.98) {0.8};

\node[text=drawColor,anchor=base,inner sep=0pt, outer sep=0pt, scale=  0.79] at (134.00, 17.98) {0.9};
\end{scope}
\begin{scope}
\path[clip] (  0.00,  0.00) rectangle (144.54,130.09);
\definecolor{drawColor}{RGB}{0,0,0}

\node[text=drawColor,anchor=base,inner sep=0pt, outer sep=0pt, scale=  0.99] at ( 83.59,  7.44) {Fraction of Outliers};
\end{scope}
\begin{scope}
\path[clip] (  0.00,  0.00) rectangle (144.54,130.09);
\definecolor{drawColor}{RGB}{0,0,0}

\node[text=drawColor,rotate= 90.00,anchor=base,inner sep=0pt, outer sep=0pt, scale=  0.99] at ( 12.32, 69.81) {Error};
\end{scope}
\begin{scope}
\path[clip] (  0.00,  0.00) rectangle (144.54,130.09);

\path[] ( 34.66, 74.70) rectangle (110.35,129.63);
\end{scope}
\begin{scope}
\path[clip] (  0.00,  0.00) rectangle (144.54,130.09);

\path[] ( 40.16, 94.66) rectangle ( 54.62,109.11);
\end{scope}
\begin{scope}
\path[clip] (  0.00,  0.00) rectangle (144.54,130.09);
\definecolor{drawColor}{RGB}{0,0,0}

\path[draw=drawColor,line width= 1.7pt,line join=round] ( 41.61,101.88) -- ( 53.17,101.88);
\end{scope}
\begin{scope}
\path[clip] (  0.00,  0.00) rectangle (144.54,130.09);
\definecolor{drawColor}{RGB}{0,0,0}

\path[draw=drawColor,line width= 1.7pt,line join=round] ( 41.61,101.88) -- ( 53.17,101.88);
\end{scope}
\begin{scope}
\path[clip] (  0.00,  0.00) rectangle (144.54,130.09);

\path[] ( 40.16, 80.20) rectangle ( 54.62, 94.66);
\end{scope}
\begin{scope}
\path[clip] (  0.00,  0.00) rectangle (144.54,130.09);
\definecolor{drawColor}{RGB}{0,0,255}

\path[draw=drawColor,line width= 1.7pt,line join=round] ( 41.61, 87.43) -- ( 53.17, 87.43);
\end{scope}
\begin{scope}
\path[clip] (  0.00,  0.00) rectangle (144.54,130.09);
\definecolor{drawColor}{RGB}{0,0,255}

\path[draw=drawColor,line width= 1.7pt,line join=round] ( 41.61, 87.43) -- ( 53.17, 87.43);
\end{scope}
\begin{scope}
\path[clip] (  0.00,  0.00) rectangle (144.54,130.09);
\definecolor{drawColor}{RGB}{0,0,0}

\node[text=drawColor,anchor=base west,inner sep=0pt, outer sep=0pt, scale=  0.99] at ( 60.12, 98.47) {Low Noise};
\end{scope}
\begin{scope}
\path[clip] (  0.00,  0.00) rectangle (144.54,130.09);
\definecolor{drawColor}{RGB}{0,0,0}

\node[text=drawColor,anchor=base west,inner sep=0pt, outer sep=0pt, scale=  0.99] at ( 60.12, 84.02) {No Noise};
\end{scope}
\begin{scope}
\path[clip] (  0.00,  0.00) rectangle (144.54,130.09);
\definecolor{drawColor}{RGB}{0,0,0}

\node[text=drawColor,anchor=base,inner sep=0pt, outer sep=0pt, scale=  1.00] at ( 83.59,117.70) {Low Noise ($\sigma = 10^{-9}$)};
\end{scope}
\end{tikzpicture} \hspace{-.25cm}
\begin{tikzpicture}[x=1pt,y=1pt]
\definecolor{fillColor}{RGB}{255,255,255}
\path[use as bounding box,fill=fillColor,fill opacity=0.00] (0,0) rectangle (144.54,130.09);
\begin{scope}
\path[clip] (  0.00,  0.00) rectangle (144.54,130.09);
\definecolor{drawColor}{RGB}{255,255,255}
\definecolor{fillColor}{RGB}{255,255,255}

\path[draw=drawColor,line width= 0.6pt,line join=round,line cap=round,fill=fillColor] (  0.00,  0.00) rectangle (144.54,130.09);
\end{scope}
\begin{scope}
\path[clip] ( 28.15, 29.36) rectangle (139.04,110.25);
\definecolor{fillColor}{gray}{0.92}

\path[fill=fillColor] ( 28.15, 29.36) rectangle (139.04,110.25);
\definecolor{drawColor}{RGB}{255,255,255}

\path[draw=drawColor,line width= 0.3pt,line join=round] ( 28.15, 43.28) --
	(139.04, 43.28);

\path[draw=drawColor,line width= 0.3pt,line join=round] ( 28.15, 63.80) --
	(139.04, 63.80);

\path[draw=drawColor,line width= 0.3pt,line join=round] ( 28.15, 84.33) --
	(139.04, 84.33);

\path[draw=drawColor,line width= 0.3pt,line join=round] ( 28.15,104.85) --
	(139.04,104.85);

\path[draw=drawColor,line width= 0.3pt,line join=round] ( 39.49, 29.36) --
	( 39.49,110.25);

\path[draw=drawColor,line width= 0.3pt,line join=round] ( 52.09, 29.36) --
	( 52.09,110.25);

\path[draw=drawColor,line width= 0.3pt,line join=round] ( 64.69, 29.36) --
	( 64.69,110.25);

\path[draw=drawColor,line width= 0.3pt,line join=round] ( 77.29, 29.36) --
	( 77.29,110.25);

\path[draw=drawColor,line width= 0.3pt,line join=round] ( 89.90, 29.36) --
	( 89.90,110.25);

\path[draw=drawColor,line width= 0.3pt,line join=round] (102.50, 29.36) --
	(102.50,110.25);

\path[draw=drawColor,line width= 0.3pt,line join=round] (115.10, 29.36) --
	(115.10,110.25);

\path[draw=drawColor,line width= 0.3pt,line join=round] (127.70, 29.36) --
	(127.70,110.25);

\path[draw=drawColor,line width= 0.6pt,line join=round] ( 28.15, 33.02) --
	(139.04, 33.02);

\path[draw=drawColor,line width= 0.6pt,line join=round] ( 28.15, 53.54) --
	(139.04, 53.54);

\path[draw=drawColor,line width= 0.6pt,line join=round] ( 28.15, 74.07) --
	(139.04, 74.07);

\path[draw=drawColor,line width= 0.6pt,line join=round] ( 28.15, 94.59) --
	(139.04, 94.59);

\path[draw=drawColor,line width= 0.6pt,line join=round] ( 33.19, 29.36) --
	( 33.19,110.25);

\path[draw=drawColor,line width= 0.6pt,line join=round] ( 45.79, 29.36) --
	( 45.79,110.25);

\path[draw=drawColor,line width= 0.6pt,line join=round] ( 58.39, 29.36) --
	( 58.39,110.25);

\path[draw=drawColor,line width= 0.6pt,line join=round] ( 70.99, 29.36) --
	( 70.99,110.25);

\path[draw=drawColor,line width= 0.6pt,line join=round] ( 83.59, 29.36) --
	( 83.59,110.25);

\path[draw=drawColor,line width= 0.6pt,line join=round] ( 96.20, 29.36) --
	( 96.20,110.25);

\path[draw=drawColor,line width= 0.6pt,line join=round] (108.80, 29.36) --
	(108.80,110.25);

\path[draw=drawColor,line width= 0.6pt,line join=round] (121.40, 29.36) --
	(121.40,110.25);

\path[draw=drawColor,line width= 0.6pt,line join=round] (134.00, 29.36) --
	(134.00,110.25);
\definecolor{drawColor}{RGB}{0,0,0}

\path[draw=drawColor,line width= 1.7pt,line join=round] ( 33.19, 33.13) --
	( 45.79, 33.19) --
	( 58.39, 33.22) --
	( 70.99, 33.43) --
	( 83.59, 33.61) --
	( 96.20, 33.80) --
	(108.80, 34.69) --
	(121.40, 41.35) --
	(134.00,106.58);
\definecolor{drawColor}{RGB}{0,0,255}

\path[draw=drawColor,line width= 1.7pt,line join=round] ( 33.19, 33.05) --
	( 45.79, 33.04) --
	( 58.39, 33.09) --
	( 70.99, 33.19) --
	( 83.59, 33.21) --
	( 96.20, 33.28) --
	(108.80, 33.46) --
	(121.40, 37.00) --
	(134.00, 86.38);
\end{scope}
\begin{scope}
\path[clip] (  0.00,  0.00) rectangle (144.54,130.09);
\definecolor{drawColor}{RGB}{190,190,190}

\node[text=drawColor,rotate= 90.00,anchor=base,inner sep=0pt, outer sep=0pt, scale=  0.62] at ( 22.23, 33.02) {0.0};

\node[text=drawColor,rotate= 90.00,anchor=base,inner sep=0pt, outer sep=0pt, scale=  0.62] at ( 22.23, 53.54) {0.5};

\node[text=drawColor,rotate= 90.00,anchor=base,inner sep=0pt, outer sep=0pt, scale=  0.62] at ( 22.23, 74.07) {1.0};

\node[text=drawColor,rotate= 90.00,anchor=base,inner sep=0pt, outer sep=0pt, scale=  0.62] at ( 22.23, 94.59) {1.5};
\end{scope}
\begin{scope}
\path[clip] (  0.00,  0.00) rectangle (144.54,130.09);
\definecolor{drawColor}{gray}{0.20}

\path[draw=drawColor,line width= 0.6pt,line join=round] ( 25.40, 33.02) --
	( 28.15, 33.02);

\path[draw=drawColor,line width= 0.6pt,line join=round] ( 25.40, 53.54) --
	( 28.15, 53.54);

\path[draw=drawColor,line width= 0.6pt,line join=round] ( 25.40, 74.07) --
	( 28.15, 74.07);

\path[draw=drawColor,line width= 0.6pt,line join=round] ( 25.40, 94.59) --
	( 28.15, 94.59);
\end{scope}
\begin{scope}
\path[clip] (  0.00,  0.00) rectangle (144.54,130.09);
\definecolor{drawColor}{gray}{0.20}

\path[draw=drawColor,line width= 0.6pt,line join=round] ( 33.19, 26.61) --
	( 33.19, 29.36);

\path[draw=drawColor,line width= 0.6pt,line join=round] ( 45.79, 26.61) --
	( 45.79, 29.36);

\path[draw=drawColor,line width= 0.6pt,line join=round] ( 58.39, 26.61) --
	( 58.39, 29.36);

\path[draw=drawColor,line width= 0.6pt,line join=round] ( 70.99, 26.61) --
	( 70.99, 29.36);

\path[draw=drawColor,line width= 0.6pt,line join=round] ( 83.59, 26.61) --
	( 83.59, 29.36);

\path[draw=drawColor,line width= 0.6pt,line join=round] ( 96.20, 26.61) --
	( 96.20, 29.36);

\path[draw=drawColor,line width= 0.6pt,line join=round] (108.80, 26.61) --
	(108.80, 29.36);

\path[draw=drawColor,line width= 0.6pt,line join=round] (121.40, 26.61) --
	(121.40, 29.36);

\path[draw=drawColor,line width= 0.6pt,line join=round] (134.00, 26.61) --
	(134.00, 29.36);
\end{scope}
\begin{scope}
\path[clip] (  0.00,  0.00) rectangle (144.54,130.09);
\definecolor{drawColor}{RGB}{190,190,190}

\node[text=drawColor,anchor=base,inner sep=0pt, outer sep=0pt, scale=  0.79] at ( 33.19, 17.98) {0.1};

\node[text=drawColor,anchor=base,inner sep=0pt, outer sep=0pt, scale=  0.79] at ( 45.79, 17.98) {0.2};

\node[text=drawColor,anchor=base,inner sep=0pt, outer sep=0pt, scale=  0.79] at ( 58.39, 17.98) {0.3};

\node[text=drawColor,anchor=base,inner sep=0pt, outer sep=0pt, scale=  0.79] at ( 70.99, 17.98) {0.4};

\node[text=drawColor,anchor=base,inner sep=0pt, outer sep=0pt, scale=  0.79] at ( 83.59, 17.98) {0.5};

\node[text=drawColor,anchor=base,inner sep=0pt, outer sep=0pt, scale=  0.79] at ( 96.20, 17.98) {0.6};

\node[text=drawColor,anchor=base,inner sep=0pt, outer sep=0pt, scale=  0.79] at (108.80, 17.98) {0.7};

\node[text=drawColor,anchor=base,inner sep=0pt, outer sep=0pt, scale=  0.79] at (121.40, 17.98) {0.8};

\node[text=drawColor,anchor=base,inner sep=0pt, outer sep=0pt, scale=  0.79] at (134.00, 17.98) {0.9};
\end{scope}
\begin{scope}
\path[clip] (  0.00,  0.00) rectangle (144.54,130.09);
\definecolor{drawColor}{RGB}{0,0,0}

\node[text=drawColor,anchor=base,inner sep=0pt, outer sep=0pt, scale=  0.99] at ( 83.59,  7.44) {Fraction of Outliers};
\end{scope}
\begin{scope}
\path[clip] (  0.00,  0.00) rectangle (144.54,130.09);

\path[] ( 25.93, 74.70) rectangle (119.08,129.63);
\end{scope}
\begin{scope}
\path[clip] (  0.00,  0.00) rectangle (144.54,130.09);

\path[] ( 31.43, 94.66) rectangle ( 45.89,109.11);
\end{scope}
\begin{scope}
\path[clip] (  0.00,  0.00) rectangle (144.54,130.09);
\definecolor{drawColor}{RGB}{0,0,0}

\path[draw=drawColor,line width= 1.7pt,line join=round] ( 32.88,101.88) -- ( 44.44,101.88);
\end{scope}
\begin{scope}
\path[clip] (  0.00,  0.00) rectangle (144.54,130.09);
\definecolor{drawColor}{RGB}{0,0,0}

\path[draw=drawColor,line width= 1.7pt,line join=round] ( 32.88,101.88) -- ( 44.44,101.88);
\end{scope}
\begin{scope}
\path[clip] (  0.00,  0.00) rectangle (144.54,130.09);

\path[] ( 31.43, 80.20) rectangle ( 45.89, 94.66);
\end{scope}
\begin{scope}
\path[clip] (  0.00,  0.00) rectangle (144.54,130.09);
\definecolor{drawColor}{RGB}{0,0,255}

\path[draw=drawColor,line width= 1.7pt,line join=round] ( 32.88, 87.43) -- ( 44.44, 87.43);
\end{scope}
\begin{scope}
\path[clip] (  0.00,  0.00) rectangle (144.54,130.09);
\definecolor{drawColor}{RGB}{0,0,255}

\path[draw=drawColor,line width= 1.7pt,line join=round] ( 32.88, 87.43) -- ( 44.44, 87.43);
\end{scope}
\begin{scope}
\path[clip] (  0.00,  0.00) rectangle (144.54,130.09);
\definecolor{drawColor}{RGB}{0,0,0}

\node[text=drawColor,anchor=base west,inner sep=0pt, outer sep=0pt, scale=  0.99] at ( 51.39, 98.47) {Medium Noise};
\end{scope}
\begin{scope}
\path[clip] (  0.00,  0.00) rectangle (144.54,130.09);
\definecolor{drawColor}{RGB}{0,0,0}

\node[text=drawColor,anchor=base west,inner sep=0pt, outer sep=0pt, scale=  0.99] at ( 51.39, 84.02) {No Noise};
\end{scope}
\begin{scope}
\path[clip] (  0.00,  0.00) rectangle (144.54,130.09);
\definecolor{drawColor}{RGB}{0,0,0}

\node[text=drawColor,anchor=base,inner sep=0pt, outer sep=0pt, scale=  1.00] at ( 83.59,117.70) {Medium Noise ($\sigma = 10^{-3}$)};
\end{scope}
\end{tikzpicture} \hspace{-.25cm}
\begin{tikzpicture}[x=1pt,y=1pt]
\definecolor{fillColor}{RGB}{255,255,255}
\path[use as bounding box,fill=fillColor,fill opacity=0.00] (0,0) rectangle (144.54,130.09);
\begin{scope}
\path[clip] (  0.00,  0.00) rectangle (144.54,130.09);
\definecolor{drawColor}{RGB}{255,255,255}
\definecolor{fillColor}{RGB}{255,255,255}

\path[draw=drawColor,line width= 0.6pt,line join=round,line cap=round,fill=fillColor] (  0.00,  0.00) rectangle (144.54,130.09);
\end{scope}
\begin{scope}
\path[clip] ( 28.15, 29.36) rectangle (139.04,110.25);
\definecolor{fillColor}{gray}{0.92}

\path[fill=fillColor] ( 28.15, 29.36) rectangle (139.04,110.25);
\definecolor{drawColor}{RGB}{255,255,255}

\path[draw=drawColor,line width= 0.3pt,line join=round] ( 28.15, 43.11) --
	(139.04, 43.11);

\path[draw=drawColor,line width= 0.3pt,line join=round] ( 28.15, 63.30) --
	(139.04, 63.30);

\path[draw=drawColor,line width= 0.3pt,line join=round] ( 28.15, 83.48) --
	(139.04, 83.48);

\path[draw=drawColor,line width= 0.3pt,line join=round] ( 28.15,103.67) --
	(139.04,103.67);

\path[draw=drawColor,line width= 0.3pt,line join=round] ( 39.49, 29.36) --
	( 39.49,110.25);

\path[draw=drawColor,line width= 0.3pt,line join=round] ( 52.09, 29.36) --
	( 52.09,110.25);

\path[draw=drawColor,line width= 0.3pt,line join=round] ( 64.69, 29.36) --
	( 64.69,110.25);

\path[draw=drawColor,line width= 0.3pt,line join=round] ( 77.29, 29.36) --
	( 77.29,110.25);

\path[draw=drawColor,line width= 0.3pt,line join=round] ( 89.90, 29.36) --
	( 89.90,110.25);

\path[draw=drawColor,line width= 0.3pt,line join=round] (102.50, 29.36) --
	(102.50,110.25);

\path[draw=drawColor,line width= 0.3pt,line join=round] (115.10, 29.36) --
	(115.10,110.25);

\path[draw=drawColor,line width= 0.3pt,line join=round] (127.70, 29.36) --
	(127.70,110.25);

\path[draw=drawColor,line width= 0.6pt,line join=round] ( 28.15, 33.02) --
	(139.04, 33.02);

\path[draw=drawColor,line width= 0.6pt,line join=round] ( 28.15, 53.20) --
	(139.04, 53.20);

\path[draw=drawColor,line width= 0.6pt,line join=round] ( 28.15, 73.39) --
	(139.04, 73.39);

\path[draw=drawColor,line width= 0.6pt,line join=round] ( 28.15, 93.58) --
	(139.04, 93.58);

\path[draw=drawColor,line width= 0.6pt,line join=round] ( 33.19, 29.36) --
	( 33.19,110.25);

\path[draw=drawColor,line width= 0.6pt,line join=round] ( 45.79, 29.36) --
	( 45.79,110.25);

\path[draw=drawColor,line width= 0.6pt,line join=round] ( 58.39, 29.36) --
	( 58.39,110.25);

\path[draw=drawColor,line width= 0.6pt,line join=round] ( 70.99, 29.36) --
	( 70.99,110.25);

\path[draw=drawColor,line width= 0.6pt,line join=round] ( 83.59, 29.36) --
	( 83.59,110.25);

\path[draw=drawColor,line width= 0.6pt,line join=round] ( 96.20, 29.36) --
	( 96.20,110.25);

\path[draw=drawColor,line width= 0.6pt,line join=round] (108.80, 29.36) --
	(108.80,110.25);

\path[draw=drawColor,line width= 0.6pt,line join=round] (121.40, 29.36) --
	(121.40,110.25);

\path[draw=drawColor,line width= 0.6pt,line join=round] (134.00, 29.36) --
	(134.00,110.25);
\definecolor{drawColor}{RGB}{0,0,0}

\path[draw=drawColor,line width= 1.7pt,line join=round] ( 33.19, 60.88) --
	( 45.79, 60.85) --
	( 58.39, 60.55) --
	( 70.99, 60.88) --
	( 83.59, 60.21) --
	( 96.20, 60.73) --
	(108.80, 67.42) --
	(121.40, 76.62) --
	(134.00,106.58);
\definecolor{drawColor}{RGB}{0,0,255}

\path[draw=drawColor,line width= 1.7pt,line join=round] ( 33.19, 33.05) --
	( 45.79, 33.04) --
	( 58.39, 33.09) --
	( 70.99, 33.19) --
	( 83.59, 33.21) --
	( 96.20, 33.28) --
	(108.80, 33.45) --
	(121.40, 36.93) --
	(134.00, 85.50);
\end{scope}
\begin{scope}
\path[clip] (  0.00,  0.00) rectangle (144.54,130.09);
\definecolor{drawColor}{RGB}{190,190,190}

\node[text=drawColor,rotate= 90.00,anchor=base,inner sep=0pt, outer sep=0pt, scale=  0.62] at ( 22.23, 33.02) {0.0};

\node[text=drawColor,rotate= 90.00,anchor=base,inner sep=0pt, outer sep=0pt, scale=  0.62] at ( 22.23, 53.20) {0.5};

\node[text=drawColor,rotate= 90.00,anchor=base,inner sep=0pt, outer sep=0pt, scale=  0.62] at ( 22.23, 73.39) {1.0};

\node[text=drawColor,rotate= 90.00,anchor=base,inner sep=0pt, outer sep=0pt, scale=  0.62] at ( 22.23, 93.58) {1.5};
\end{scope}
\begin{scope}
\path[clip] (  0.00,  0.00) rectangle (144.54,130.09);
\definecolor{drawColor}{gray}{0.20}

\path[draw=drawColor,line width= 0.6pt,line join=round] ( 25.40, 33.02) --
	( 28.15, 33.02);

\path[draw=drawColor,line width= 0.6pt,line join=round] ( 25.40, 53.20) --
	( 28.15, 53.20);

\path[draw=drawColor,line width= 0.6pt,line join=round] ( 25.40, 73.39) --
	( 28.15, 73.39);

\path[draw=drawColor,line width= 0.6pt,line join=round] ( 25.40, 93.58) --
	( 28.15, 93.58);
\end{scope}
\begin{scope}
\path[clip] (  0.00,  0.00) rectangle (144.54,130.09);
\definecolor{drawColor}{gray}{0.20}

\path[draw=drawColor,line width= 0.6pt,line join=round] ( 33.19, 26.61) --
	( 33.19, 29.36);

\path[draw=drawColor,line width= 0.6pt,line join=round] ( 45.79, 26.61) --
	( 45.79, 29.36);

\path[draw=drawColor,line width= 0.6pt,line join=round] ( 58.39, 26.61) --
	( 58.39, 29.36);

\path[draw=drawColor,line width= 0.6pt,line join=round] ( 70.99, 26.61) --
	( 70.99, 29.36);

\path[draw=drawColor,line width= 0.6pt,line join=round] ( 83.59, 26.61) --
	( 83.59, 29.36);

\path[draw=drawColor,line width= 0.6pt,line join=round] ( 96.20, 26.61) --
	( 96.20, 29.36);

\path[draw=drawColor,line width= 0.6pt,line join=round] (108.80, 26.61) --
	(108.80, 29.36);

\path[draw=drawColor,line width= 0.6pt,line join=round] (121.40, 26.61) --
	(121.40, 29.36);

\path[draw=drawColor,line width= 0.6pt,line join=round] (134.00, 26.61) --
	(134.00, 29.36);
\end{scope}
\begin{scope}
\path[clip] (  0.00,  0.00) rectangle (144.54,130.09);
\definecolor{drawColor}{RGB}{190,190,190}

\node[text=drawColor,anchor=base,inner sep=0pt, outer sep=0pt, scale=  0.79] at ( 33.19, 17.98) {0.1};

\node[text=drawColor,anchor=base,inner sep=0pt, outer sep=0pt, scale=  0.79] at ( 45.79, 17.98) {0.2};

\node[text=drawColor,anchor=base,inner sep=0pt, outer sep=0pt, scale=  0.79] at ( 58.39, 17.98) {0.3};

\node[text=drawColor,anchor=base,inner sep=0pt, outer sep=0pt, scale=  0.79] at ( 70.99, 17.98) {0.4};

\node[text=drawColor,anchor=base,inner sep=0pt, outer sep=0pt, scale=  0.79] at ( 83.59, 17.98) {0.5};

\node[text=drawColor,anchor=base,inner sep=0pt, outer sep=0pt, scale=  0.79] at ( 96.20, 17.98) {0.6};

\node[text=drawColor,anchor=base,inner sep=0pt, outer sep=0pt, scale=  0.79] at (108.80, 17.98) {0.7};

\node[text=drawColor,anchor=base,inner sep=0pt, outer sep=0pt, scale=  0.79] at (121.40, 17.98) {0.8};

\node[text=drawColor,anchor=base,inner sep=0pt, outer sep=0pt, scale=  0.79] at (134.00, 17.98) {0.9};
\end{scope}
\begin{scope}
\path[clip] (  0.00,  0.00) rectangle (144.54,130.09);
\definecolor{drawColor}{RGB}{0,0,0}

\node[text=drawColor,anchor=base,inner sep=0pt, outer sep=0pt, scale=  0.99] at ( 83.59,  7.44) {Fraction of Outliers};
\end{scope}
\begin{scope}
\path[clip] (  0.00,  0.00) rectangle (144.54,130.09);

\path[] ( 33.36, 74.70) rectangle (111.65,129.63);
\end{scope}
\begin{scope}
\path[clip] (  0.00,  0.00) rectangle (144.54,130.09);

\path[] ( 38.86, 94.66) rectangle ( 53.31,109.11);
\end{scope}
\begin{scope}
\path[clip] (  0.00,  0.00) rectangle (144.54,130.09);
\definecolor{drawColor}{RGB}{0,0,0}

\path[draw=drawColor,line width= 1.7pt,line join=round] ( 40.30,101.88) -- ( 51.87,101.88);
\end{scope}
\begin{scope}
\path[clip] (  0.00,  0.00) rectangle (144.54,130.09);
\definecolor{drawColor}{RGB}{0,0,0}

\path[draw=drawColor,line width= 1.7pt,line join=round] ( 40.30,101.88) -- ( 51.87,101.88);
\end{scope}
\begin{scope}
\path[clip] (  0.00,  0.00) rectangle (144.54,130.09);

\path[] ( 38.86, 80.20) rectangle ( 53.31, 94.66);
\end{scope}
\begin{scope}
\path[clip] (  0.00,  0.00) rectangle (144.54,130.09);
\definecolor{drawColor}{RGB}{0,0,255}

\path[draw=drawColor,line width= 1.7pt,line join=round] ( 40.30, 87.43) -- ( 51.87, 87.43);
\end{scope}
\begin{scope}
\path[clip] (  0.00,  0.00) rectangle (144.54,130.09);
\definecolor{drawColor}{RGB}{0,0,255}

\path[draw=drawColor,line width= 1.7pt,line join=round] ( 40.30, 87.43) -- ( 51.87, 87.43);
\end{scope}
\begin{scope}
\path[clip] (  0.00,  0.00) rectangle (144.54,130.09);
\definecolor{drawColor}{RGB}{0,0,0}

\node[text=drawColor,anchor=base west,inner sep=0pt, outer sep=0pt, scale=  0.99] at ( 58.81, 98.47) {High Noise};
\end{scope}
\begin{scope}
\path[clip] (  0.00,  0.00) rectangle (144.54,130.09);
\definecolor{drawColor}{RGB}{0,0,0}

\node[text=drawColor,anchor=base west,inner sep=0pt, outer sep=0pt, scale=  0.99] at ( 58.81, 84.02) {No Noise};
\end{scope}
\begin{scope}
\path[clip] (  0.00,  0.00) rectangle (144.54,130.09);
\definecolor{drawColor}{RGB}{0,0,0}

\node[text=drawColor,anchor=base,inner sep=0pt, outer sep=0pt, scale=  1.00] at ( 83.59,117.70) {High Noise ($\sigma = 10^{-1}$)};
\end{scope}
\end{tikzpicture}
\hspace{-2cm}
\caption{GLIMPS performs satisfactorily under low and medium noise conditions. However, high noise conditions considerably affect its performance. In this experiment, 30\% entries of $\x$ were removed in the first stage.}
\label{noise}
\end{figure*}
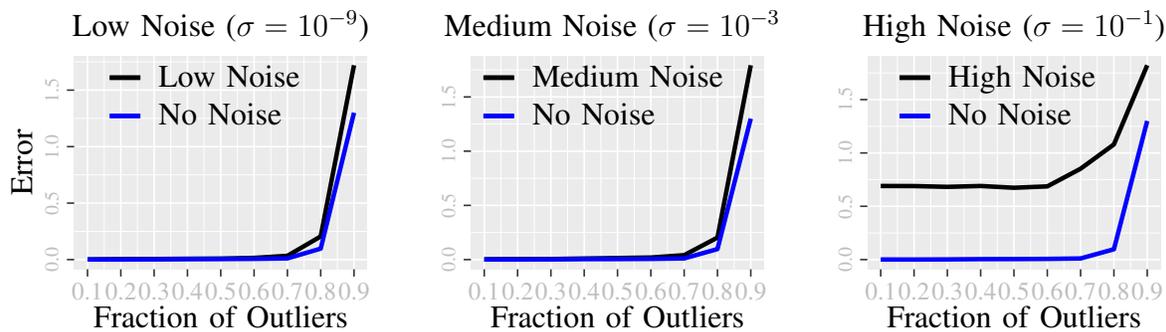

Finally, we study the behavior of GLIMPS under different noise conditions: \emph{low, medium} and \emph{high}. We use three values of sigma $\sigma=1e^{-9}$ (low), $\sigma=1e^{-3}$ (medium), and $\sigma=1e^{-1}$ (high) for three levels of noises. We observe that GLIMPS steadily tolerates up to medium level of noise. However, high noise severely affects GLIMPS performance as shown in Fig. \ref{noise}. 

\section{Conclusion}
In this paper, a two-stage approach (GLIMPS) for super robust matched subspace detection is proposed to improve the performance of MILP formulation in terms of both accuracy and computation time. In the first stage, a greedy algorithm is applied to remove some percentage of outliers and thus we obtain a reduced input vector. Later on, we feed the reduced input vector and warm-start MILP to increase scalability and performance of the same MILP. It is observed that GLIMPS achieves better results over {\em Erasure} (greedy algorithm) and MILP with full accuracy for fraction of outliers up to 80\%, whereas, greedy algorithm and MILP individually can tolerate up to 60\% and 76\% outliers respectively. We also show that if we use greedy+$\ell_1$-minimization, it can't achieve higher accuracy than only greedy algorithm. In case of GLIMPS, we notice that if outliers are abundant, greedy algorithm erroneously removes some inliers as outliers in the first stage that, in turn, restricts GLIMPS to succeed beyond 80\%. Our next focus is to find necessary and sufficient deterministic theoretic conditions that will guarantee no error to be incurred in greedy algorithm (first stage) so that GLIMPS can tolerate more outliers.

\end{document}